\documentclass[10pt,twocolumn,letterpaper]{article}

\usepackage{cvpr}
\usepackage{times}
\usepackage{epsfig}
\usepackage{graphicx}
\usepackage{amsmath}
\usepackage{amssymb}

\usepackage{url}            
\usepackage{booktabs}       
\usepackage{amsfonts}       
\usepackage{nicefrac}       
\usepackage{microtype}      
\usepackage{multirow}
\usepackage{threeparttable}
\usepackage{caption}
\usepackage{subfigure}
\usepackage{wrapfig}
\usepackage{color}
\usepackage{diagbox}

\usepackage{appendix}

\usepackage[breaklinks=true,bookmarks=false]{hyperref}

\cvprfinalcopy 

\def\cvprPaperID{****} 

\begin{document}

\title{Competitive Inner-Imaging Squeeze and Excitation for Residual Network}

\author{Yang Hu,
        Guihua Wen \thanks{corresponding author},
        Mingnan Luo \thanks{equal contribution with Guihua Wen},
        Dan Dai,
        Jiajiong Ma,
        Zhiwen Yu\\
School of Computer Science \& Engineering, South China University of Technology\\
Panyu, Guangzhou, Guangdong, China\\
\small
\{cssuperhy@mail., crghwen@, csluomingnan@mail., csdaidan@mail., csmajiajiong@mail., zhwyu@\}scut.edu.cn
}

\maketitle

\begin{abstract}
Residual networks, which use a residual unit to supplement the identity mappings, enable very deep convolutional architecture to operate well, however, the residual architecture has been proved to be diverse and redundant, which may leads to low-efficient modeling. In this work, we propose a competitive squeeze-excitation (SE) mechanism for the residual network. Re-scaling the value for each channel in this structure will be determined by the residual and identity mappings jointly, and this design enables us to expand the meaning of channel relationship modeling in residual blocks. Modeling of the competition between residual and identity mappings cause the identity flow to control the complement of the residual feature maps for itself. Furthermore, we design a novel inner-imaging competitive SE block to shrink the consumption and re-image the global features of intermediate network structure, by using the inner-imaging mechanism, we can model the channel-wise relations with convolution in spatial. We carry out experiments on the CIFAR, SVHN, and ImageNet datasets, and the proposed method can challenge state-of-the-art results.
\end{abstract}

\section{Introduction}
\begin{figure*}[!ht]
\centering
\includegraphics[scale=0.7]{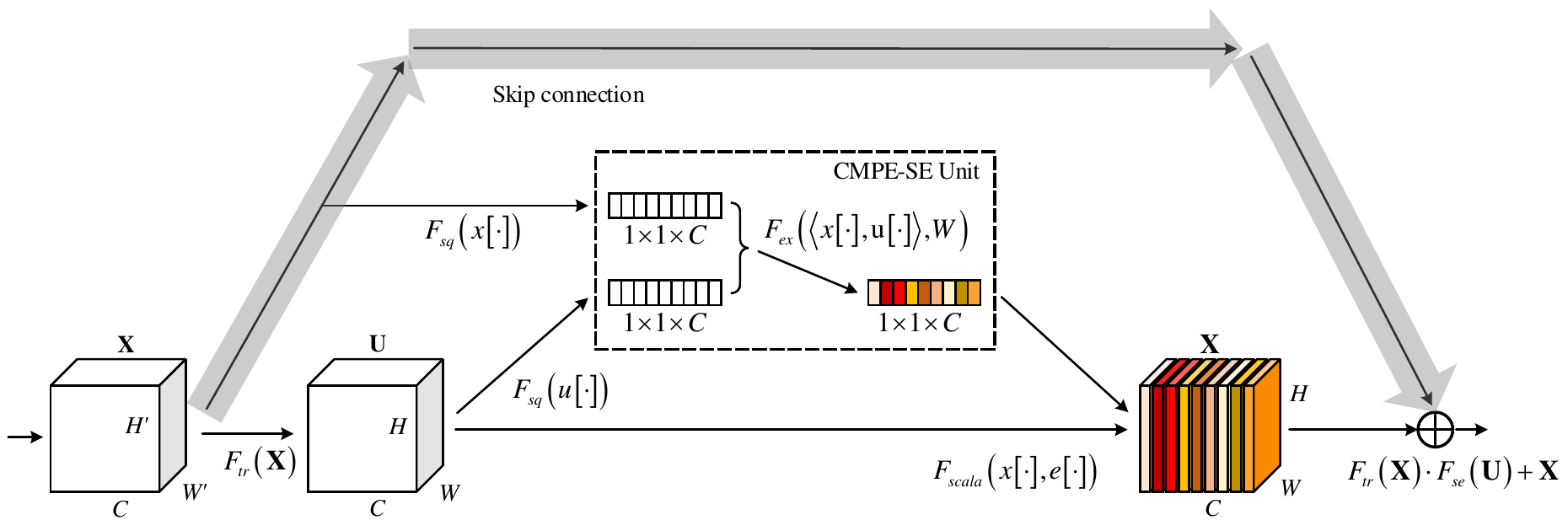}
\caption{Competitive Squeeze-Excitation Architecture for Residual block.}
\label{fig1}
\end{figure*}

Deep convolutional neural networks (CNNs) have exhibited significant effectiveness in tackling and modeling image data~\cite{krizhevsky2012,szegedy2015,szegedy2016,simonyan2015}. The presentation of the residual network (ResNet) enables the network structure go far deeper and achieve superior performance~\cite{he2016}. Moreover, attention has also been paid to the modeling of implicit relationships in CNNs~\cite{chen2017,wang2017residual}. The "\textit{squeeze-excitation}" (SE-Net) architecture~\cite{hu2018} captures the channel relationships with a low cost, and can be used directly in all CNN types. However, when a SE-block is applied in ResNet, the identity mapping does not take into account the input of the channel-wise attention of the residual flow. For analysis of ResNet, the residual mapping can be regarded as a supplement to the identical mapping~\cite{he2016Identity}, and with the increase in depth, the residual network exhibits a certain amount of redundancy~\cite{huang2016,veit2016residual}; thus, identity mappings should also consider channel attention, thereby making the supplement for itself more dynamic and precise, under the known condition that the residual network has extremely high redundancy.

In this work, we design a new, competitive squeeze and excitation architecture based on the SE-block, known as the competitive SE (CMPE-SE) network. We aim to expand the factors considered in the channel re-weighting of residual mappings and use the CMPE-SE design to model the implicit competitive relationship between identity and residual feature maps. Furthermore, we attempt to presents a novel strategy to alleviate the redundancy of ResNets with the CMPE-SE mechanism, it makes residual mappings tend to provide more efficient supplementary for identity mappings.

Compared to the typical SE building block, the composition of the CMPE-SE block is illustrated in Fig. \ref{fig1}. The basic mode of the CMPE-SE module absorbs the compressed signals for identity mappings $\mathrm{X} \in \mathbb{R}^{W' \times H' \times C'}$ and residual mappings $\mathrm{U} \in \mathbb{R}^{W \times H \times C}$, and with the same squeeze operation as in reference~\cite{hu2018}, concatenates and embeds these jointly and multiplies the excitation value back to each channel. Moreover, the global distributions from residual and identity feature maps can be stitched into new relational maps, we call this operation as "Inner-Imaging". Through "Inner-Imaging", we can use convolution filters to model the relationships between channels in spatial location, and various filters can be tested on the inner-imaged maps.

As the design of the CMPE-SE module considers residual and identity flow jointly, based on the original SE block for ResNet, it expands the task and meaning of "\textit{squeeze and excitation}", recalibrating the channel-wise features. The modeling object of the CMPE-SE unit is not limited to the relationship of the residual channels, but the relationship between all residual and identity feature maps, as well as the competition between residual and identity flows. In this manner, the network can dynamically adjust the complementary weights of residual channels to the identity mapping by using the competitive relations in each residual block. Furthermore, "Inner-Imaging" enable us to encode the channel-wise relationship with convolution filters, at the same time, it also provide diversified and spatial internal representation for the architecture of ResNet.

The exploration of convolutional network architecture and modeling of network internal representation is a meaningful and challenging task~\cite{xie2017genetic,zoph2018learning}, typically with high complicacy~\cite{zhang2017polynet,wang2018multi-scale}. In comparison, the layout of the CMPE-SE module outlined above is easy to implement and can be cheaply applied to the residual network and its all variants. The contributions of this study can be listed as follows.

\begin{itemize}
\item We present a new strategy to alleviate the redundancy of residual network and enhance its modeling efficiency, with the novel competitive "\textit{squeeze and excitation}" unit, which jointly models the relationship of residual and identity channels, the identity mapping can participate in the re-weighting for residual channels.
\item We propose a inner-imaging design for intermediate structure representation in CNNs, in order to re-scan the channel relation features with convolutional filters. Furthermore, we try to fold the re-imaged channel relation maps and explore more possibilities of convolutional channel relationship encoder.
\item We conduct experiments on several datasets, including CIFAR-10, CIFAR-100, SVHN, and ImageNet, to validate the performance of the presented models. Moreover, we discover our approach can stimulate the potential of the smaller networks.
\end{itemize}

\section{Related Work}
\begin{figure*}[!ht]
\centering
\includegraphics[scale=0.48]{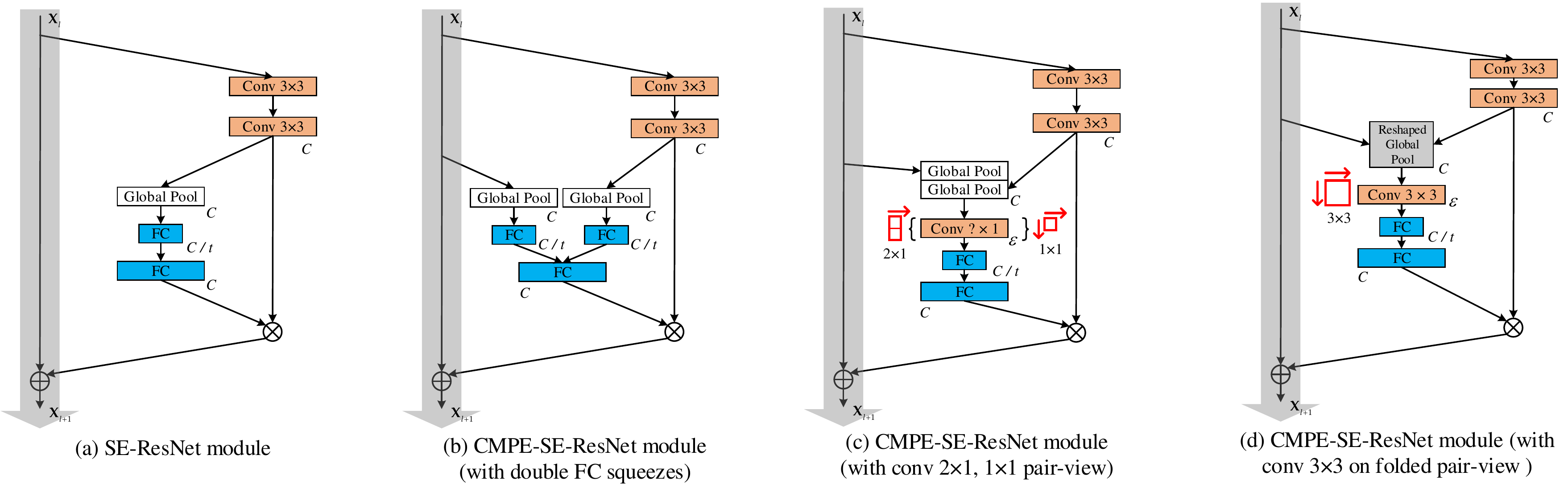}
\caption{Difference between SE-ResNet module and CMPE-SE-ResNet modules. (a) Typical SE residual block: the orange rectangles represent the convolution, the blue indicates the fully connected layer, and the white is the global average pooling. (b) CMPE-SE residual block of the version with double fully connected embedding for squeezed signals, which are merged in the excitation layer. (c) CMPE-SE residual block with $2 \times 1$ or $1 \times 1$ convolutional pair-view; after stacking the squeezed signals, the red pane and arrow indicate the size and scanning direction of the convolution. (d) CMPE-SE residual block with $3 \times 3$ convolutional pair-view after folding the squeezed signal maps.}
\label{fig2}
\end{figure*}

\textbf{Residual architectures.} ResNet~\cite{he2016} has become popular by virtue of its assistance in deep model training. Numerous works based thereon improve performance by expanding its structure~\cite{zagoruyko2016wide,han2017deep,xie2017aggregated,zhang2017polynet} or use its explanation of ordinary differential equations to explore its reversible form~\cite{chang2018reversible,chen2018neural}. Because ResNet is internally diverse without operations such as the "drop-path"~\cite{larsson2017fractalnet} and has been proven to be structurally redundant~\cite{veit2016residual}, destructive approaches may promote its efficiency and enrich the structural representation by means of policy learning~\cite{wu2018blockdrop,wang2017skipnet} or a dynamic exit strategy~\cite{figurnov2017spatially,huang2017multi-scale}.

A parallel line of research has deemed that intermediate feature maps should be modeled repeatedly~\cite{huang2017densely}. This compact architecture enables intermediate features to be refined and expanded, thereby enhancing the representation ability with concentrated parameter sizes. ~\cite{yang2018convolutional} proposed a more compact model by circulating the dense block. Furthermore, dual-path networks (DPNs)~\cite{chen2017dual} combine the advantages of ResNet and DenseNet, and cause the residual units to perform extra modeling of the relationship between the identity and densely connected flow. A trend of compact architectures is to expand the mission of the network subassemblies while refining the intermediate features. Based on the SE block~\cite{hu2018}, our proposed CMPE-SE design also refines the intermediate features and develops the role of the SE unit. The difference is that our model focuses on self-controlling of components in ResNet, rather than simple feature reuse. Moreover, the re-imaging of channel signals presents a novel modeling view of intermediate features.

\textbf{Attention and gating mechanisms in CNNs.} Attention is widely applied in the modeling process of CNNs~\cite{nguyen2018attentive} and is typically used to re-weight the image spatial signals~\cite{wang2017residual,li2018harmonious,zheng2017learning,Sun2018Multi}, including multi-scale~\cite{chen2016attention,newell2016stacked} and multi-shape~\cite{jaderberg2015spatial} features. As a tool for biasing the allocation of resources~\cite{hu2018}, attention is also used to regulate the internal CNN features of~\cite{perez2018film,stollenga2014deep}. Unlike channel switching, combination~\cite{zhang2017interleaved,zhang2018shufflenet} or using reinforcement learning to reorganize the network paths~\cite{Ahmed2017Connectivity}, channel-wise attention, typically such as~\cite{hu2018}, provides an end-to-end training solution for re-weighting the intermediate channel features. Moreover, certain models combine spatial and channel-wise attention~\cite{chen2017,Linsley2018Global,Sanghyun2018CBAM}, and their modeling scope is still limited in total attentional elements. In contrast, our proposed CMPE-SE block considers the additional related factors (identity mappings) apart from the objects of attention (residual mappings). Furthermore, we test the effects of various convolutional filters in channel-wise attention with channel signal inner-imaging, which can mine the spatial channel-wise relations.

\section{Competitive Squeeze Excitation Blocks}
\noindent The residual block is routinely defined as the amalgamation of identity mapping $\mathrm{X} \in \mathbb{R}^{W' \times H' \times C'}$ and residual mapping $\mathrm{U} \in \mathbb{R}^{W \times H \times C}$, as follows:
\begin{equation}\label{eq1}
y = F_{res}(\mathrm{x}, \mathrm{w}_{r}) + \mathrm{x}.
\end{equation}

We record the output of the residual mapping as $\mathrm{U}_{r}=F_{res}(\mathrm{x}, \mathrm{w}_{r})= [\mathrm{u}_{r}^{1}, \mathrm{u}_{r}^{2}, \dots, \mathrm{u}_{r}^{C}]$. As described in the design of SE-Net, the "\textit{squeeze-excitation}" module~\cite{hu2018} controls the re-weighted value of the convolution feature maps including the residual mappings, as follows:
\begin{equation}\label{eq2}
\hat{u}_{r}^{c} = F_{sq}(\mathrm{u}_{r}^{c}) = \frac{1}{W \times H} \sum_{i=1}^{W}\sum_{j=1}^{H}u_{r}^{c}(i, j),
\end{equation}
\begin{equation}\label{eq3}
\mathrm{s} = F_{ex}(\hat{\mathrm{u}}_{r},\mathrm{w}_{ex}) = \sigma \left(\mathit{ReLU}(\hat{\mathrm{u}}_{r}, \mathrm{w}_{1}), \mathrm{w}_{2} \right),
\end{equation}
\begin{equation}\label{eq4}
\begin{aligned}
\tilde{\mathrm{x}}_{c} &= F_{scale}( s_{c}, \mathrm{u}_{r}^{c}) \\
&= F_{se}(\mathrm{u}_{r})[\cdot] \times F_{res}(\mathrm{x}, \mathrm{w}_{r})[\cdot] = s_{c} \cdot \mathrm{u}_{r}^{c},
\end{aligned}
\end{equation}
where $\hat{u}_{r}^{c}$ refers to the global pooling result of the squeeze operation, $\sigma(\cdot)$ denotes the sigmoid activation, and operators $\times$ and $\cdot$ are the element-wise multiplication. The excitation contains two fully connected (FC) layers, the weights $\mathrm{w}_{1} \in \mathbb{R}^{\frac{C}{t} \times C}$ mean dimensionality-reduction with the ratio $t$ (set to $16$ by default) and $\mathrm{w}_{2} \in \mathbb{R}^{C \times \frac{C}{t}}$, so the variable $\mathrm{s}$ is the rescaling tensor for the residual channels. We can summarize the flow of the residual block in SE-ResNet as:
\begin{equation}\label{eq5}
y = F_{se}(\mathrm{u}_{r}) \cdot F_{res}(\mathrm{x}, \mathrm{w}_{r}) + \mathrm{x}.
\end{equation}

Stated thus, the conventional SE operation models the relationship of the convolution channels and feedback by recalibrating values that are calculated only using the feature maps of the residual flow in ResNet.

\subsection{Competition between Residual and Identity Flows}
\noindent The architecture of the current SE-ResNet illustrates that the rebuilding weights are not products of the joint decision with identity and residual mappings. From an intuitional point of view, we introduce the identity flow into the process of "\textit{squeeze-excitation}".

Corresponding to the residual mapping $\mathrm{U}_{r}$, the global information embedding from the identity mapping $\mathrm{X}_{id} = [\mathrm{x}_{id}^{1}, \mathrm{x}_{id}^{2}, \dots, \mathrm{x}_{id}^{C}]$ can also be obtained as:
\begin{equation}\label{eq6}
\hat{x}_{id}^{c} = F_{sq}(\mathrm{x}_{id}^{c}) = \frac{1}{W \times H} \sum_{i=1}^{W}\sum_{j=1}^{H}{x}_{id}^{c}(i, j),
\end{equation}
and as with $\hat{u}_{r}^{c}$, $\hat{x}_{id}^{c}$ is the global average pooling of identity features, and is used as a part of the joint input for the residual channel recalibration, together with $\hat{u}_{r}^{c}$:
\begin{equation}\label{eq7}
\begin{aligned}
\mathrm{s} &= F_{ex}(\hat{\mathrm{u}}_{r}, \hat{\mathrm{x}}_{id}, \mathrm{w}_{ex}) \\
&= \sigma \left(\langle \mathit{ReLU}(\hat{\mathrm{u}}_{r}, \mathrm{w}_{1}^{r}), \mathit{ReLU}(\hat{\mathrm{x}}_{id}, \mathrm{w}_{1}^{id}) \rangle, \mathrm{w}_{2}^{ex} \right),
\end{aligned}
\end{equation}
\begin{equation}\label{eq8}
\tilde{\mathrm{x}}_{c} = F_{se}(\mathrm{u}_{r}, \mathrm{x}_{id})[\cdot] \times F_{res}(\mathrm{x}_{id}, \mathrm{w}_{r})[\cdot] = s_{c} \cdot \mathrm{u}_{r}^{c},
\end{equation}
where the parameters $\mathrm{w}_{1}^{r} \in \mathbb{R}^{\frac{C}{t} \times C}$ and $\mathrm{w}_{1}^{id} \in \mathbb{R}^{\frac{C}{t} \times C}$ encode the squeezed signals from the identity and residual mappings, and are followed by another FC layer parameterized by $\mathrm{w}_{2}^{ex} \in \mathbb{R}^{C \times \frac{2C}{t}}$, with $C$ neurons.

The competition between the residual and identity mappings is modeled by the CMPE-SE module introduced above, and reacts to each residual channel. Implicitly, we can believe that the winning of the identity channels in this competition results in less weights of the residual channels, while the weights of the residual channels will increase. Finally, the CMPE-SE residual block is reformulated as:
\begin{equation}\label{eq9}
y = F_{se}(\mathrm{u}_{r}, \mathrm{x}_{id}) \cdot F_{res}(\mathrm{x}_{id}, \mathrm{w}_{r}) + \mathrm{x}_{id}.
\end{equation}

Figures \ref{fig2}(a) and (b) illustrate the difference between the typical SE and CMPE-SE residual modules. The embedding of the squeezed signals $\hat{\mathrm{u}}_{r} = [\hat{u}_{r}^{1}, \hat{u}_{r}^{2}, \dots, \hat{u}_{r}^{C} ]^{\top}$ and $\hat{\mathrm{x}}_{id} = [\hat{x}_{id}^{1}, \hat{x}_{id}^{2}, \dots, \hat{x}_{id}^{C} ]^{\top}$ are simply concatenated prior to excitation. Here, the back-propagation algorithm optimizes two intertwined parts of modeling processes: (1) the relationships of all channels in the residual block; and (2) the competition between the residual and identity channels. Moreover, $\mathrm{w}_{1}^{id}$ is the only additional parameter cost.

\subsection{Pair-View Re-imaging for Intermediate Channel Features}
\begin{figure}[!ht]
\centering
\includegraphics[scale=0.65]{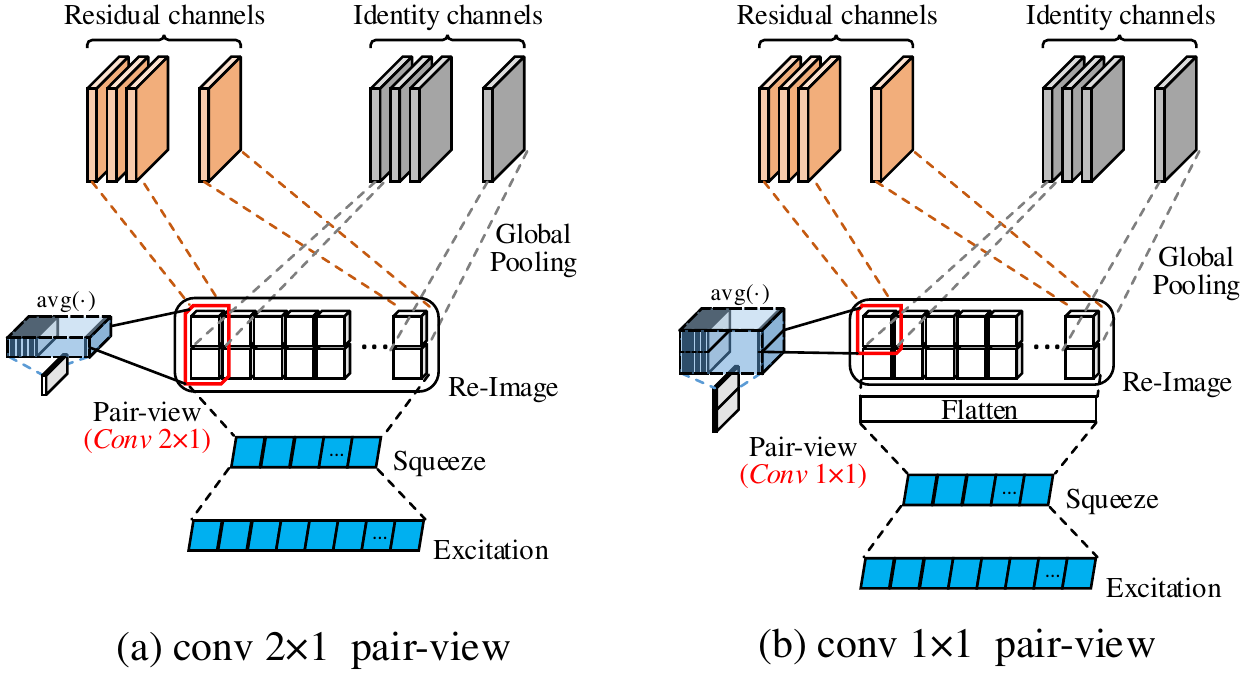}
\caption{Pair-view modes of competitive squeeze and excitation with convolution $2 \times 1$ and $1 \times 1$.}
\label{fig3}
\end{figure}

\noindent In the basic mode of the CMPE-SE residual block, one additional FC encoder is required for joint modeling of the competition of the residual and identity channels. We also design the pair-view strategies of the competitive "\textit{Squeeze-Excitation}" to save parameters and capture the channel relation features from a novel angle. Figures \ref{fig2}(c) illustrate their structures.

Firstly, the stacked squeezed feature maps are generated as:
\begin{equation}\label{eq10}
\hat{\mathrm{v}}_{s} = \left[
  \begin{matrix}
   \hat{\mathrm{u}}_{r}^{\top} \\
   \hat{\mathrm{x}}_{id}^{\top}
  \end{matrix}
  \right] = \left[
 \begin{matrix}
   \hat{u}_{r}^{1}, & \hat{u}_{r}^{2} & \cdots & \hat{u}_{r}^{C} \\
   \hat{x}_{id}^{1}, & \hat{x}_{id}^{2} & \cdots & \hat{x}_{id}^{C}
  \end{matrix}
  \right],
\end{equation}
where the inner-imaging encoder acquires the feature maps of the channel relations rather than the original picture input. We use $\varepsilon$ filters $\{\mathrm{w}_{(2 \times 1)}^{1}, \dots, \mathrm{w}_{(2 \times 1)}^{\varepsilon} \}, \ \mathrm{w}_{(2 \times 1)}^{i} = [w_{11}^{i}, w_{21}^{i}]^{\top}$ scan the stacked tensor of squeezed features from the residual and identity channels, and then average the pair-view outputs,
\begin{equation}\label{eq11}
\mathrm{v}_{c} = \frac{1}{\varepsilon}\sum_{i=1}^{\varepsilon}\left(\hat{\mathrm{v}}_{s} \ast \mathrm{w}_{(2 \times 1)}^{i} \right)^{\top},
\end{equation}
where $\ast$ denotes the convolution and $\mathrm{v}_{c}$ is the re-imaged feature map. Batch normalization (BN)~\cite{ioffe2015batch} is performed directly following convolution. Next, re-imaged signal encoding and excitation take place, as follows:
\begin{equation}\label{eq12}
\mathrm{s} = \sigma \left(\mathrm{w}_{2}^{ex} \cdot ReLU(\mathrm{w}_{1} \cdot \mathrm{v}_{c}) \right),
\end{equation}
where the squeeze encoder is parameterized by $\mathrm{w}_{1} \in \mathbb{R}^{\frac{C}{t} \times C}$ and the excitation parameters are also shrunk to $\mathrm{w}_{2}^{ex} \in \mathbb{R}^{C \times \frac{C}{t}}$. Figure \ref{fig3}(a) illustrates the detailed structure of the "Conv $(2 \times 1)$" pair-view CMPE-SE unit.

The "Conv $(2 \times 1)$" pair-view strategy models the competition between the residual and identity channels based on strict upper and lower positions, which ignores the factor that any feature signal in the re-imaged tensor could be associated with any other signal, not only in the location of the vertical direction. Based on this consideration, we use a $1 \times 1$ convolution kernel $\mathrm{w}_{(1 \times 1)}^{i} = [w_{11}^{i}]$ to replace the above $\mathrm{w}_{(2 \times 1)}^{i} = [w_{11}^{i}, w_{21}^{i}]^{\top}$. Furthermore, a flattened layer is used to reshape the output of the $1 \times 1$ convolution:
\begin{equation}\label{eq13}
\mathrm{v}_{c}' = \frac{1}{\varepsilon}\sum_{i=1}^{\varepsilon}\left(\hat{\mathrm{v}}_{s} \ast \mathrm{w}_{(1 \times 1)}^{i} \right),
\end{equation}
\begin{equation}\label{eq14}
\mathrm{s} = \sigma \left(\mathrm{w}_{2}^{ex} \cdot ReLU \left(\mathrm{w}_{1} \cdot \left(F_{flatten} \left(\mathrm{v}_{c}' \right) \right)^{\top} \right) \right),
\end{equation}
where $\mathrm{v}_{c} = \left(F_{flatten} \left(\mathrm{v}_{c}' \right) \right)^{\top}$ corresponds to Eq. \ref{eq11}, the parameter size of the encoder will return to $\mathrm{w}_{1} \in \mathbb{R}^{\frac{C}{t} \times 2C}$, and the excitation remains $\mathrm{w}_{2}^{ex} \in \mathbb{R}^{C \times \frac{C}{t}}$. Figure \ref{fig3}(b) depicts the "Conv $(1 \times 1)$" pair-view CMPE-SE unit. In fact, this mode can be regarded as a simple linear transformation for combined squeezed signals prior to embedding. The number of pair-view convolution kernels $\varepsilon$ mentioned previously is set as the block width divided by the dimensionality-reduction ratio $t$.

\subsection{Exploration of Folded Shape for Pair-View Inner-imaging}
\begin{figure}[!ht]
\centering
\includegraphics[scale=0.8]{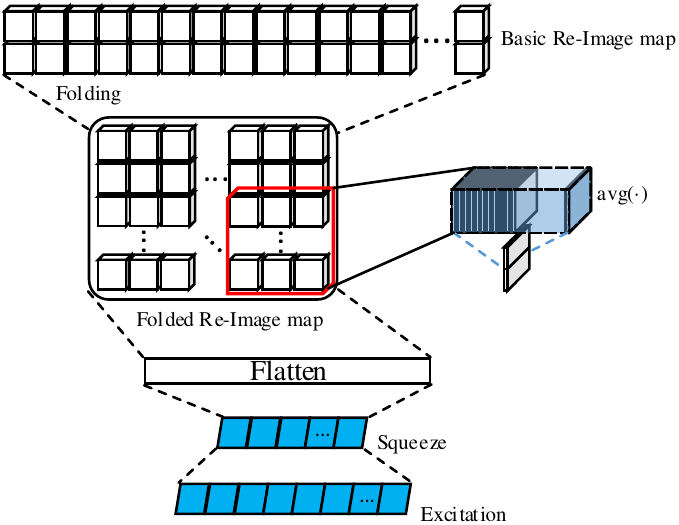}
\caption{Folded pair-view mode of competitive squeeze and excitation with convolution $3 \times 3$.}
\label{fig4}
\end{figure}

The inner-imaging design provide two shapes of convolutional kernel: "conv $(2 \times 1)$" and "conv $(1 \times 1)$" can be regard as a simple linear transformation for combined squeezed signals prior to embedding. However, too flat inner-imaged maps obstruct the diversity of filter shapes, and it is impossible to model location relationships of squeezed signals in larger fields.

In order to expand the shape of inner-imaging convolution, and provide more robust and precise channel relation modeling, we fold the pair-view re-imaged maps into more square matrices with shape of $(n \times m)$ while maintaining the alternating arrangement of squeezed signals from residual and identity channels, as follows:
\begin{equation}\label{eq15}
\begin{aligned}
\hat{\mathrm{v}}_{f} &= T \left(\hat{\mathrm{v}}_{s} \right) = \left[
\begin{matrix}
  \hat{v}_{s}^{11}, & \cdots & \hat{v}_{s}^{1m} \\
  \vdots & \ddots & \vdots \\
  \hat{v}_{s}^{n1}, & \cdots & \hat{v}_{s}^{nm}
\end{matrix}
\right] = T \left(\left[
  \begin{matrix}
   \hat{\mathrm{u}}_{r}^{\top} \\
   \hat{\mathrm{x}}_{id}^{\top}
  \end{matrix}
  \right] \right) \\
&= \left[
  \begin{matrix}
    \hat{u}_{r}^{1}, & \hat{u}_{r}^{2} & \cdots & \hat{u}_{r}^{m} \\
    \hat{x}_{id}^{1}, & \hat{x}_{id}^{2} & \cdots & \hat{x}_{id}^{m} \\
    \hat{u}_{r}^{m+1}, & \hat{u}_{r}^{m+2} & \cdots & \hat{u}_{r}^{2 \cdot m} \\
    \hat{x}_{id}^{m+1}, & \hat{x}_{id}^{m+2} & \cdots & \hat{x}_{id}^{2 \cdot m} \\
    \vdots & \vdots & \ddots & \vdots \\
    \hat{u}_{r}^{C-m+1}, & \hat{u}_{r}^{C-m+2} & \cdots & \hat{u}_{r}^{C} \\
    \hat{x}_{id}^{C-m+1}, & \hat{x}_{id}^{C-m+2} & \cdots & \hat{x}_{id}^{C}
  \end{matrix}
  \right],
\end{aligned}
\end{equation}
where $T(\cdot)$ is the reshape function to fold the basic inner-imaged maps and we receive the folded matrix $\hat{\mathrm{v}}_{f}$.

Then, we can freely expand the shape of inner-imaging convolution kernel to $3 \times 3$ as $\mathrm{w}_{3 \times 3}^{i}$, and use it scan the folded pair-view maps as follows, the structure details of folded pair-view are also shown in Figures \ref{fig2}(d) and \ref{fig4}.
\begin{equation}\label{eq16}
\mathrm{v}_{c}' = \frac{1}{\varepsilon}\sum_{i=1}^{\varepsilon} \left(\hat{\mathrm{v}}_{f} \ast \mathrm{w}_{(3 \times 3)}^{i} \right)
\end{equation}

Acquiescently, in folded mode of pair-view encoders, the flatten layer is used to reshape the convolution results for subsequent FC layers, as $\mathrm{v}_{c} = \left(F_{flatten} \left(\mathrm{v}_{c}' \right) \right)^{\top}$.

To sum up, the proposed CMPE-SE mechanism can technically improve the efficiency of residual network modeling through the following two characteristics: 1. Directly participating by identity flow, in the re-weighting of residual channels, makes the complementary modeling more efficient; 2. The mechanism of inner-imaging and its folded mode explore the richer forms of channel relationship modeling.

\section{Experiments}
\noindent We evaluate our approach on the \textbf{CIFAR-10, CIFAR-100, SVHN and ImageNet} datasets. We train several basic ResNets and compare their performances with/without the CMPE-SE module. Thereafter, we challenge the state-of-the-art results.

\subsection{Datasets and Settings}
\noindent \textbf{CIFAR.} The CIFAR-10 and CIFAR-100 datasets consist of $32 \times 32$ colored images~\cite{Krizhevsky2009Learning}. Both datasets contain 60,000 images belonging to 10 and 100 classes, with 50,000 images for training and 10,000 images for testing. We subtract the mean and divide by the standard deviation for data normalization, and standard data augmentation (translation/mirroring) is adopted for the training sets.

\noindent \textbf{SVHN.} The Street View House Number (SVHN) dataset~\cite{Netzer2011Reading} contains $32 \times 32$ colored images of 73,257 samples in the training set and 26,032 for testing, with 531,131 digits for additional training. We divide the images by 255 and use all training data without data augmentation.

\noindent \textbf{ImageNet.} The ILSVRC 2012 dataset~\cite{deng2009imagenet} contains $1.2$ million training images, 50,000 validation images, and 100,000 for testing, with 1,000 classes. Standard data augmentation is adopted for the training set and the $224 \times 224$ crop is randomly sampled. All images are normalized into $[0, 1]$, with mean values and standard deviations.

\noindent \textbf{Settings.} We test the effectiveness of the CMPE-SE modules on two classical models: pre-act ResNet~\cite{he2016Identity} and the Wide Residual Network~\cite{zagoruyko2016wide} with CIFAR-10 and CIFAR-100, and we also re-implement the typical SE block~\cite{hu2018} based on these. For fair comparison, we follow the basic structures and hyper-parameter turning in the original papers; further implementation details are available on the open source\footnote{https://github.com/scut-aitcm/CompetitiveSENet}. We train our models by means of optimizer stochastic gradient descent with $0.9$ Nesterov momentum, and use a batch size of 128 for 200 epochs. The learning rate is initialized to $0.1$ and divided by 10 at the 100th and 150th epochs for the pre-act ResNet, and divided by 5 at epochs 60, 120, and 160 for WRN. The \textit{mixup} is an advanced training strategy on convex combinations of sample pairs and their labels~\cite{zhang2018mixup}. We apply this to the aforementioned evaluations and add 20 epochs with the traditional strategy following the formal training process of \textit{mixup}. On the SVHN, our models are trained for 160 epochs; the initial learning rate is $0.01$, and is divided by 10 at the 80th and 120th epochs. On ImageNet, we train our models for 100 epochs with a batch size of 64. The initial learning rate is $0.1$ and it is reduced by 10 times at epochs 30, 60, and 90.

Based on experimental experience, the shape of folded re-imaging maps is set as: $(n={2C}/{16}, m=16)$ for pre-act ResNet and $(n=20, m={C}/{10})$ for WRN. In fact, the setting of this hyper-parameter does not cause serious disturbance to classification accuracy of our models.

\subsection{Results on CIFAR and SVHN}
\noindent The results of the contrast experiments for ResNets with/without the CMPE-SE module are illustrated in Tables \ref{table1} and \ref{table2}. We use the pre-act ResNet~\cite{he2016Identity} by default, where the numbers of parameters are recorded in brackets and the optimal records are marked in bold. By analyzing these results, we can draw the following conclusions:

\begin{table}[!t]
\small
\centering
\begin{tabular}{p{4.5cm}cc}
\toprule
\centering{Model (\# parames)} & C10 & C10 \textit{mixup} \\
\midrule
ResNet-110 \small{(1.7M)} & 6.37 & -- \\
ResNet-164 \small{(1.7M)} & 5.46 & 4.15 \\
\midrule
SE-ResNet-110 \small{(1.75M)} & 5.68 & -- \\
SE-ResNet-164 \small{(1.95M)} & 4.85 & 4.07 \\
\midrule
CMPE-SE-ResNet-110 (Ours)\\
\quad -- \textit{$1 \times 1$ pair-view} \small{(1.76M)} & 5.45 & 4.30 \\
CMPE-SE-ResNet-164 (Ours)\\
\quad -- \textit{Double FC} \small{(2.12M)} & 4.72 & 3.82 \\
\quad -- \textit{$2 \times 1$ pair-view} \small{(1.95M)} & 4.59 & 3.76 \\
\quad -- \textit{$1 \times 1$ pair-view} \small{(2.04M)} & \textbf{4.57} & 3.78 \\
\quad -- \textit{$3 \times 3$ folded pair-view} \small{(1.99M)} & 4.60 & \textbf{3.65} \\
\hline
\hline
 & C100 & C100 \textit{mixup} \\
\midrule
ResNet-110 \small{(1.7M)} & -- & 23.98 \\
ResNet-164 \small{(1.7M)} & 24.33 & 20.84 \\
\midrule
SE-ResNet-110 \small{(1.75M)} & 25.82 & -- \\
SE-ResNet-164 \small{(1.95M)} & 22.61 & 19.89 \\
\midrule
CMPE-SE-ResNet-110 (Ours)\\
\quad -- \textit{$1 \times 1$ pair-view} \small{(1.76M)} & 25.35 & 22.92 \\
CMPE-SE-ResNet-164 (Ours)\\
\quad -- \textit{Double FC} \small{(2.12M)} & 22.38 & 19.58 \\
\quad -- \textit{$2 \times 1$ pair-view} \small{(1.95M)} & 22.41 & 19.69 \\
\quad -- \textit{$1 \times 1$ pair-view} \small{(2.04M)} & \textbf{22.35} & 19.46 \\
\quad -- \textit{$3 \times 3$ folded pair-view} \small{(1.99M)} & 22.38 & \textbf{18.98} \\
\bottomrule
\end{tabular}
\caption{Error rates($\%$) of pre-act ResNets on CIFAR-10 and CIFAR-100 datasets.}
\label{table1}
\end{table}

\begin{table}[!t]
\small
\centering
\begin{tabular}{p{4.5cm}cc}
\toprule
\centering{Model (\# parames)} & C10 & C10 \textit{mixup} \\
\midrule
WRN-22-10 \small{(26.8M)} & 4.44 & -- \\
WRN-28-10 \small{(36.5M)} & 4.17 & 2.70 \\
\midrule
SE-WRN-22-10 \small{(27.0M)} & 4.09 & -- \\
SE-WRN-28-10 \small{(36.8M)} & 3.88 & 2.68 \\
\midrule
CMPE-SE-WRN-16-8 (Ours)\\
\quad -- \textit{$1 \times 1$ pair-view} \small{(11.1M)} & 4.20 & 3.18 \\
\quad -- \textit{$3 \times 3$ folded pair-view} \small{(11.1M)} & 4.22 & 3.18 \\
CMPE-SE-WRN-22-10 (Ours)\\
\quad -- \textit{$1 \times 1$ pair-view} \small{(27.1M)} & 3.75 & 2.86 \\
\quad -- \textit{$3 \times 3$ folded pair-view} \small{(27.1M)} & 3.78 & 2.81 \\
CMPE-SE-WRN-28-10 (Ours)\\
\quad -- \textit{Double FC} \small{(37.0M)} & 3.66 & 2.62 \\
\quad -- \textit{$2 \times 1$ pair-view} \small{(36.8M)} & 3.73 & 2.65 \\
\quad -- \textit{$1 \times 1$ pair-view} \small{(36.9M)} & \textbf{3.58} & 2.58 \\
\quad -- \textit{$3 \times 3$ folded pair-view} \small{(36.9M)} & 3.59 & \textbf{2.57} \\
\hline
\hline
 & C100 & C100 \textit{mixup} \\
\midrule
WRN-22-10 \small{(26.8M)} & 20.75 & 17.88 \\
WRN-28-10 \small{(36.5M)} & 20.50 & 17.50 \\
\midrule
SE-WRN-22-10 \small{(27.0M)} & 19.52 & 17.06 \\
SE-WRN-28-10 \small{(36.8M)} & 19.05 & 16.77 \\
\midrule
CMPE-SE-WRN-16-8 (Ours)\\
\quad -- \textit{$1 \times 1$ pair-view} \small{(11.1M)} & 19.77 & 17.26 \\
\quad -- \textit{$3 \times 3$ folded pair-view} \small{(11.1M)} & 19.40 & 17.24 \\
CMPE-SE-WRN-22-10 (Ours)\\
\quad -- \textit{$1 \times 1$ pair-view} \small{(27.1M)} & 18.86 & 16.82 \\
\quad -- \textit{$3 \times 3$ folded pair-view} \small{(27.1M)} & 18.82 & 16.52 \\
CMPE-SE-WRN-28-10 (Ours)\\
\quad -- \textit{Double FC} \small{(37.0M)} & 18.69 & 16.23 \\
\quad -- \textit{$2 \times 1$ pair-view} \small{(36.8M)} & 18.71 & 16.18 \\
\quad -- \textit{$1 \times 1$ pair-view} \small{(36.9M)} & 18.55 & 16.13 \\
\quad -- \textit{$3 \times 3$ folded pair-view} \small{(36.9M)} & \textbf{18.47} & \textbf{16.07} \\
\bottomrule
\end{tabular}
\caption{Error rates($\%$) of Wide Residual Network on CIFAR-10 and CIFAR-100 datasets.}
\label{table2}
\end{table}

The CMPE-SE block can achieve superior performance over the SE block for both the classical and wide ResNets. It reduces the error rate of SE-ResNet by $0.226\%$ on average and $0.312\%$ for WRN, and does not consume excessive extra parameters ($0.2\% \sim 5\%$ over the SE residual network). The pair-view mode of the CMPE-SE units with $1\times1$ convolution can achieve superior results over the basic mode and use less parameters, which means that hybrid modeling of squeezed signals is more effective than merging them after embedding. Another phenomenon is that the CMPE-SE module can reduce the error rate more efficaciously on the WRN model than on the traditional ResNet; therefore, the fewer number of layers and wider residual in the "dumpy" wide ResNet can better reflect the role of identity mapping in the residual channel-wise attention.

By observing the performances of ResNets under different scales, the CMPE-SE unit enables smaller networks to achieve or even exceed the same structure with additional parameters. For WRN, the classification results of the CMPE-SE-WRN-16-8 are the same as or exceed those of WRN-28-10, and the results of the CMPE-SE-WRN-22-10 are superior to those of the SE-WRN-28-10. The folded mode of CMPE-SE unit with $3 \times 3$ filters can achieve fairly or even better results than $1 \times 1$ pair-view CMPE-SE, with less parameters.

The \textit{mixup}~\cite{zhang2018mixup} can be considered as an advanced approach to data augmentation, which can improve the generalization ability of models. In the case of using the \textit{mixup}, the CMPE-SE block can further improve the performance of the residual networks until achieving state-of-the-art results.

\begin{table*}[!t]
\small
\centering
\begin{tabular}{p{6cm}ccccccc}
\toprule
\centering{Model} & Depth & \# parames & C10 & C10 \textit{mixup} & C100 & C100 \textit{mixup} & SVHN \\
\midrule
Original ResNet & 110 & 1.7M & 6.43 & -- & 25.16 & -- & -- \\
Pre-act ResNet-18  & 18 & 11.7M & -- & 4.20 & -- & 21.10 & -- \\
Stochastic depth & 110 & 1.7M & 5.23 & -- & 24.58 & -- & 1.75 \\
FractalNet & 21 & 38.6M & 4.60 & -- & 23.73 & -- & 1.87 \\
\midrule
DenseNet & 100 & 27.2M & 3.74 & -- & 19.25 & -- & 1.59 \\
DenseNet-BC & 190 & 25.6M & {\color{red} \textbf{3.46}} & 2.70 & 17.18 & 16.80 & -- \\
ResNeXt-29 & 29 & 34.4M & 3.65 & -- & 17.77 & -- & -- \\
PyramidNet ($\alpha$ = 270) & 110 & 28.3M & 3.73 & -- & 18.25 & -- & -- \\
\quad -- \textit{bottleneck} & 164 & 27.0M & 3.48 & -- & {\color{red} \textbf{17.01}} & -- & -- \\
CliqueNet-30 & 30 & 10.02M & 5.06 & -- & 21.83 & -- & 1.64 \\
\midrule
CMPE-SE-WRN-28-10 (Ours)\\
\quad -- \textit{double FC} & 28 & 37.04M & 3.66 & 2.62 & 18.69 & 16.23 & 1.61 \\
\quad -- \textit{$1 \times 1$ pair-view} & 28 & 36.92M & \textbf{3.58} & 2.58 & 18.55 & 16.13 & 1.59 \\
\quad -- \textit{$3 \times 3$ folded pair-view} & 28 & 36.90M & 3.59 & {\color{red} \textbf{2.57}} & \textbf{18.47} & {\color{red} \textbf{16.07}} & {\color{red} \textbf{1.59}} \\
\bottomrule
\end{tabular}
\caption{Error rates ($\%$) of different methods on CIFAR-10, CIFAR-100, and SVHN datasets. The best records of our models are in \textbf{bold} and the best results are highlighted in {\color{red} \textbf{red}}.}
\label{table3}
\end{table*}

Table \ref{table3} lists the challenge results of the CMPE-SE-WRN-28-10 with state-of-the-art results. The compared networks include: original ResNet~\cite{he2016}, pre-act ResNet~\cite{he2016Identity}, ResNet with stochastic depth~\cite{huang2016}, FractalNet~\cite{larsson2017fractalnet}, DenseNet~\cite{huang2017densely}, ResNeXt~\cite{xie2017aggregated}, PyramidNet~\cite{han2017deep}, and CliqueNet~\cite{yang2018convolutional}. We observe that our models based on wide residual networks can achieve comparable or superior performance to the compared models. Moreover, we know that although the parameter size taken is large, the training speed of the WRN is significantly faster than DenseNets, and even faster than ResNets~\cite{zagoruyko2016wide}. Considering the high extensibility of proposed CMPE-SE mechanism on all ResNet variants, it is reasonable to believe that the CMPE-SE module can achieve better results on some more complex residual achitectures.

\subsection{Results on ImageNet}
\begin{table}[!ht]
\small
\centering
\begin{tabular}{p{4.5cm}cc}
\toprule
\centering{Model} & top-1 & top-5 \\
\midrule
ResNet-18 & 30.43 & 10.76 \\
ResNet-34 & 26.73 & 8.74 \\
ResNet-50 & 24.01 & 7.02 \\
DenseNet-121 & 25.02 & 7.71 \\
CliqueNet & 24.98 & 7.48 \\
\midrule
SE-ResNet-50 & 23.29 & 6.62 \\
SE-CliqueNet & 24.01 & 7.15 \\
\midrule
CMPE-SE-ResNet-50 (Ours)\\
\quad -- \textit{Double FC} & 23.06 & 6.46 \\
\quad -- \textit{$1 \times 1$ pair-view} & 22.97 & 6.41 \\
\quad -- \textit{$3 \times 3$ folded pair-view} & \textbf{22.79} & \textbf{6.35} \\
\bottomrule
\end{tabular}
\caption{Single crop error rates ($\%$) on ImageNet.}
\label{table4}
\end{table}

\noindent Owing to the limitation of computational resources (GTX 1080Ti $\times$ 2), we only test the performance of the pre-act ResNet-50 (ImageNet mode) after being equipped with CMPE-SE blocks, and we use the smaller mini-batch with a size of 64, instead of 256 as in most studies.

Although a smaller batch size would impair the performance training for the same epochs~\cite{yang2018convolutional}, the results of the CMPE-SE-ResNet-50 (both double FC and $1 \times 1$ pair-view modes) are slightly superior to those of other models at the same level, such as SE-ResNet-50~\cite{hu2018}. Compared to the SE-ResNet-50, the CMPE-SE-ResNet-50 with $3 \times 3$ folded inner-imaging can reduce the top-1 error rate by $0.5\%$ and the top-5 error rate by $0.27\%$. The other compared models contain the pre-act ResNet-18, 34, and 50~\cite{he2016Identity}, DenseNet-121~\cite{huang2017densely}, CliqueNet, and SE-CliqueNet~\cite{yang2018convolutional}, where "SE-CliqueNet" means CliqueNet uses channel-wise attentional transition.

\subsection{Discussion}
\begin{figure*}[!t]
\centering
\includegraphics[scale=0.6]{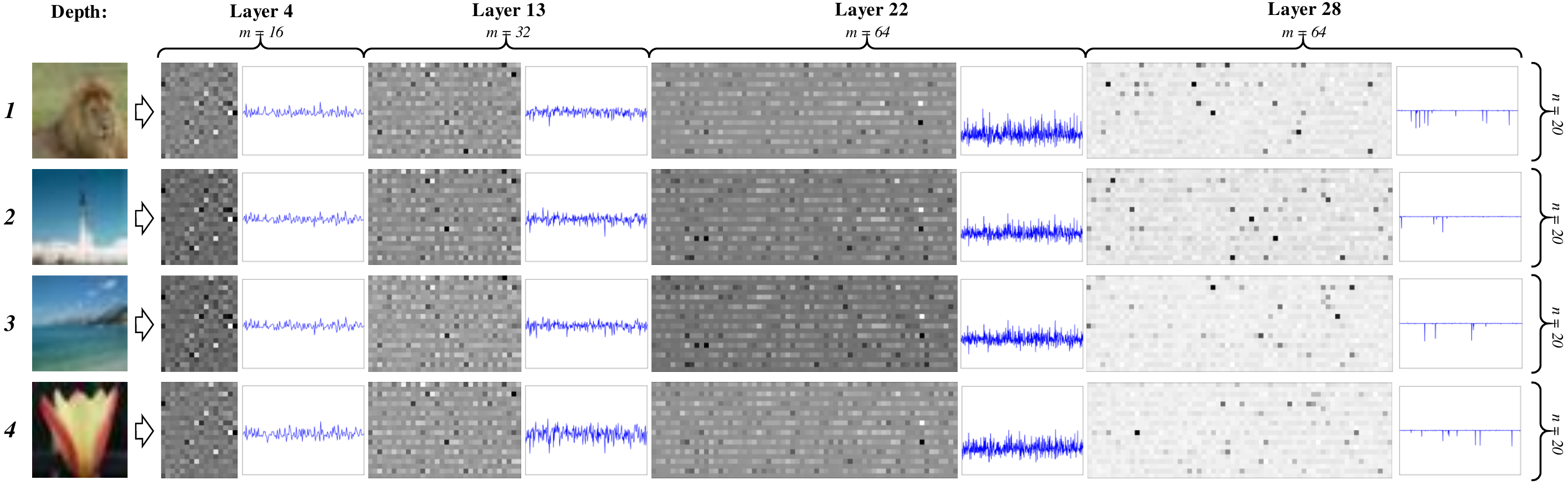}
\caption{Examples of folded inner-imaged maps and attentional values from model \textbf{CMPE-SE-WRN-28-10} on different samples in \textbf{CIFAR-100}, which are from layer 4, 13, 22, 28, and the shape of inner-imaged map is $(n \times m)$.}
\label{fig5}
\end{figure*}

\noindent Compared to the promotion of the CMPE-SE module on the same level networks, another fact is highly noteworthy: our CMPE-SE unit can greatly stimulate the potential of smaller networks with fewer parameters, enabling them to achieve or even exceed the performance of larger models. This improvement proves that the refined modeling for the inner features in convolutional networks is necessary.

Regarding the refined modeling of intermediate convolutional features, DenseNet~\cite{huang2017densely} is a type of robust repetitive refinement for inner feature maps, while the "squeeze and excitation"~\cite{hu2018} can also be considered as a type of refined modeling for channel features, and its refining task is learning the relationships of convolution channels. Furthermore, the CMPE-SE module extends the task of refined modeling for intermediate features. So that we can make the modeling process of ResNets more efficient.

In addition to modeling the competitive relationship between the residual and identity mappings, the CMPE-SE module also provides the fundamental environment for re-imaging the intermediate residual and identity features. In order to facilitate the display, Fig. \ref{fig5} illustrates several examples of fragmented inner-imaging parts and the corresponding excitation outputs. These re-imaged maps come from different layers in depths 4, 13, 22, and 28, and we can observe that the average pooled signal maps of different samples are largely identical with only minor differences at first, then become more diversified after multi-times attentional re-scaling. The attentional outputs show great diversity and tend to suppress the redundant residual modeling at deeper layers, until in the last layers, when the network feature maps themselves are more expressive and sparse, the attention values become stable, only with very few jumps.

Although the folded inner-imaging mechanism does not show very significant superiority over the ordinary pair-view CMPE-SE module, such a design still provides more possibilities for channel-wise squeezed signal organization and encoding, it has a strong enlightenment.

In order to reduce the parameter cost generated by the subsequent FC layers, we average the outputs of the pair-view convolution kernels. When attempting not to do so, we find that the former can save numerous parameters without sacrificing too much performance. This indicates that the inner-imaging of the channel features is parameter efficient, and we can even use a tiny and fixed number of filters to complete pair-view re-imaging.

In the study of this paper, we have only applied $2 \times 1$ and $1 \times 1$ two types of pair-view filters, and $3 \times 3$ kernels on folded pair-view encoder, which can achieve the aforementioned results. More forms of convolutional channel relation encoders can be easily added into the CMPSE-SE framework, it shows that the CMPE-SE module has high extensibility and wide application value. Also, we have reason to believe that branch competitive modeling and inner-imaging can result in more capacious re-imaged feature maps and a diverse refined modeling structure on multi-branch networks.

\section{Conclusion}
In this paper, we have presented a competitive squeeze and excitation block for ResNets, which models the competitive relation from both the residual and identity channels, and expand the task of channel-wise attentional modeling. Furthermore, we introduce the inner-imaging strategy to explore the channel relationships by convolution on re-imaged feature maps, then we fold the inner-imaged maps to enrich the channel relation encoding strategies. The proposed design uses several additional parameters and can easily be applied to any type of residual network. We evaluated our models on three publicly available datasets against the state-of-the-art results. Our approach can improve the performance of ResNets and stimulate the potential of smaller networks. Moreover, the presented method is extremely scalable and offers the potential to play a greater role in multi-branch architectures.

{\small
\bibliographystyle{ieee}
\bibliography{egbib}
}

\appendixpage
\appendices
\section{Evaluation for Different Shapes of Folded Inner-Imaging}
\begin{table*}[t]
\small
\centering
\begin{tabular}{p{4cm}cccccc}
\toprule
\centering{\multirow{2}{*}{\diagbox{Model}{Shape}}} & \multicolumn{2}{c}{$(n={2C}/{8}, m=8)$} & \multicolumn{2}{c}{$(n={2C}/{16}, m=16)$} & \multicolumn{2}{c}{$(n={2C}/{32}, m=32)$} \\
\cline{2-7}
 & CIFAR-10 & CIFAR-100 & CIFAR-10 & CIFAR-100 & CIFAR-10 & CIFAR-100 \\
\midrule
CMPE-SE-ResNet-164 & 4.78 & 22.49 & \underline{\textit{4.60}} & \underline{\textit{22.38}} & 4.68 & 22.42 \\
\midrule
\centering{\multirow{2}{*}{\diagbox{Model}{Shape}}} & \multicolumn{2}{c}{$(n=32, m={C}/{16})$} & \multicolumn{2}{c}{$(n=20, m={C}/{10})$} & \multicolumn{2}{c}{$(n=16, m={C}/{8})$} \\
\cline{2-7}
 & CIFAR-10 & CIFAR-100 & CIFAR-10 & CIFAR-100 & CIFAR-10 & CIFAR-100 \\
\midrule
CMPE-SE-WRN-16-8 & 4.28 & 19.55 & \underline{\textit{4.22}} & \underline{\textit{19.40}} & 4.24 & 19.41 \\
CMPE-SE-WRN-22-10 & 3.86 & 18.89 & \underline{\textit{3.78}} & \underline{\textit{18.82}} & 3.80 & 18.84 \\
CMPE-SE-WRN-28-10 & 3.71 & 18.53 & \underline{\textit{3.59}} & \underline{\textit{18.47}} & 3.60 & 18.52 \\
\bottomrule
\end{tabular}
\caption{Test error(\%) comparison for folded inner-imaging with different shapes on CIFAR-10 and CIFAR-100, the folded inner-imaging mechanism uses \textit{Conv 3$\times$3} encoder and the italics with underline indicate the shape we choosed.}
\label{apx_table1}
\end{table*}

Table \ref{apx_table1} lists the test error with \textit{Conv 3$\times$3} encoders in different shapes of folded inner-imaging. We can find out that the classification ability of out models is not very sensitive to the folded shape. Although the more square folded shapes can get better results, the worst results in Table \ref{apx_table1} can still achieve at the same level as the basic CMPE-SE module or better, so, some bad settings of the folded shape will not cause the performance of our models to drop drastically.

\section{Folded Inner-Imaging Examples and The Corresponding Excitation Outputs}
\begin{figure*}[t]
\centering
\includegraphics[scale=0.6]{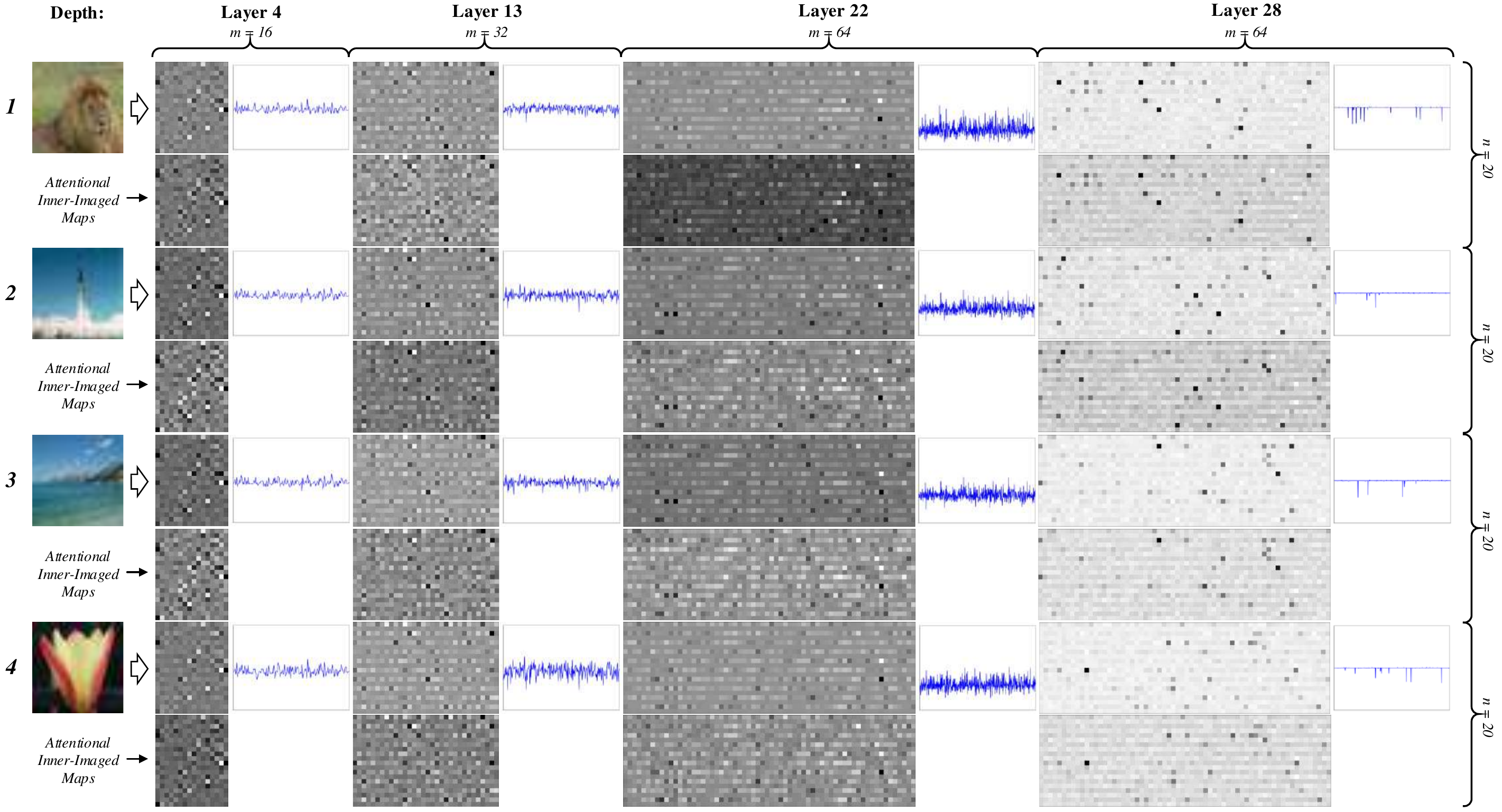}
\caption{Examples of inner-imaged maps and excitation outputs on model: \textbf{CMPE-SE-WRN-28-10}, the first line of each sample represents the initial inner-imaged maps and the next line refers the simulated inner-imaged maps after been re-weighted.}
\label{apx_fig1}
\end{figure*}
\begin{figure*}[t]
\centering
\includegraphics[scale=0.6]{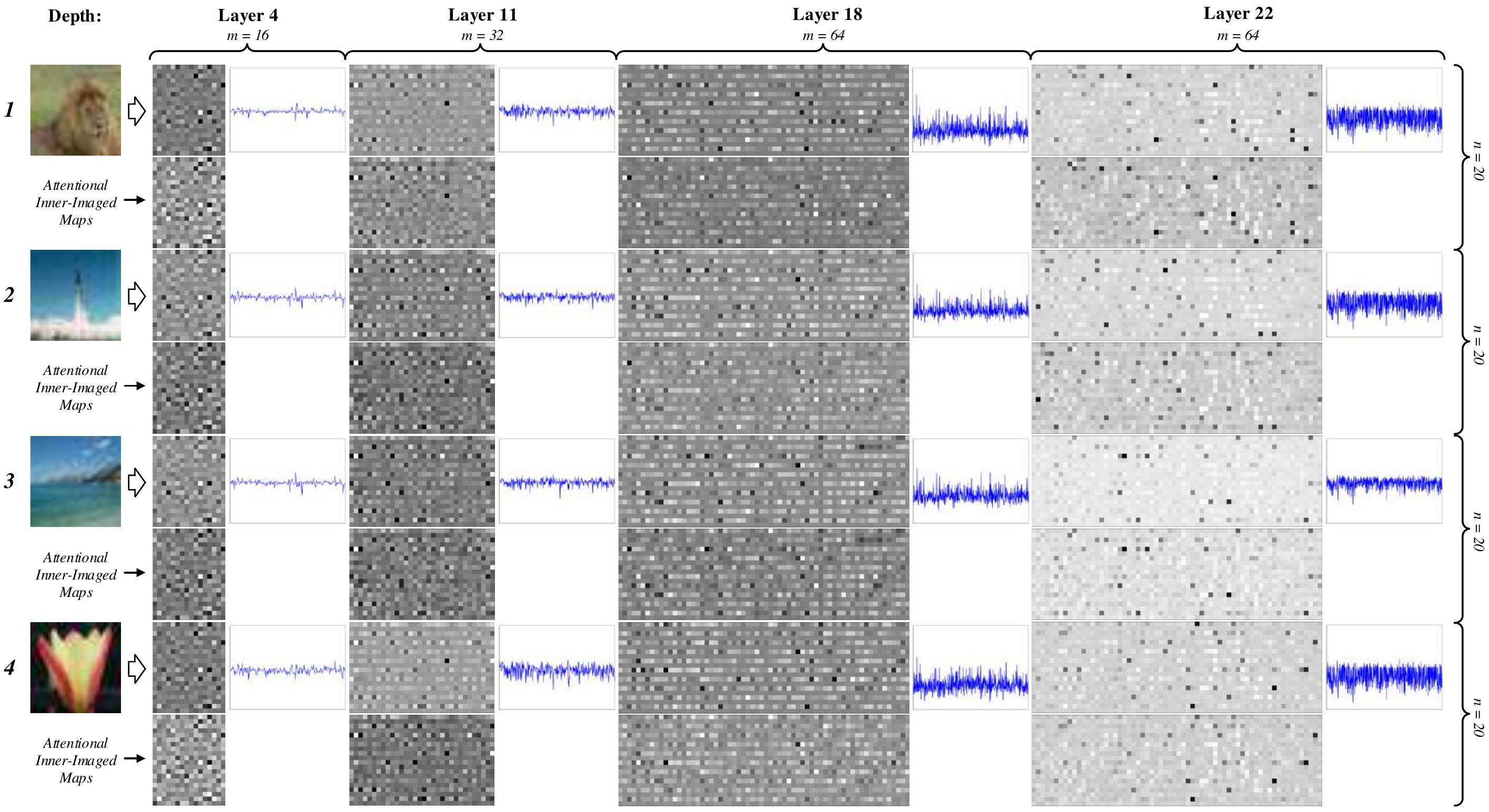}
\caption{Examples of inner-imaged maps and excitation outputs on model: \textbf{CMPE-SE-WRN-22-10}, the first line of each sample represents the initial inner-imaged maps and the next line refers the simulated inner-imaged maps after been re-weighted.}
\label{apx_fig2}
\end{figure*}
\begin{figure*}[t]
\centering
\includegraphics[scale=0.6]{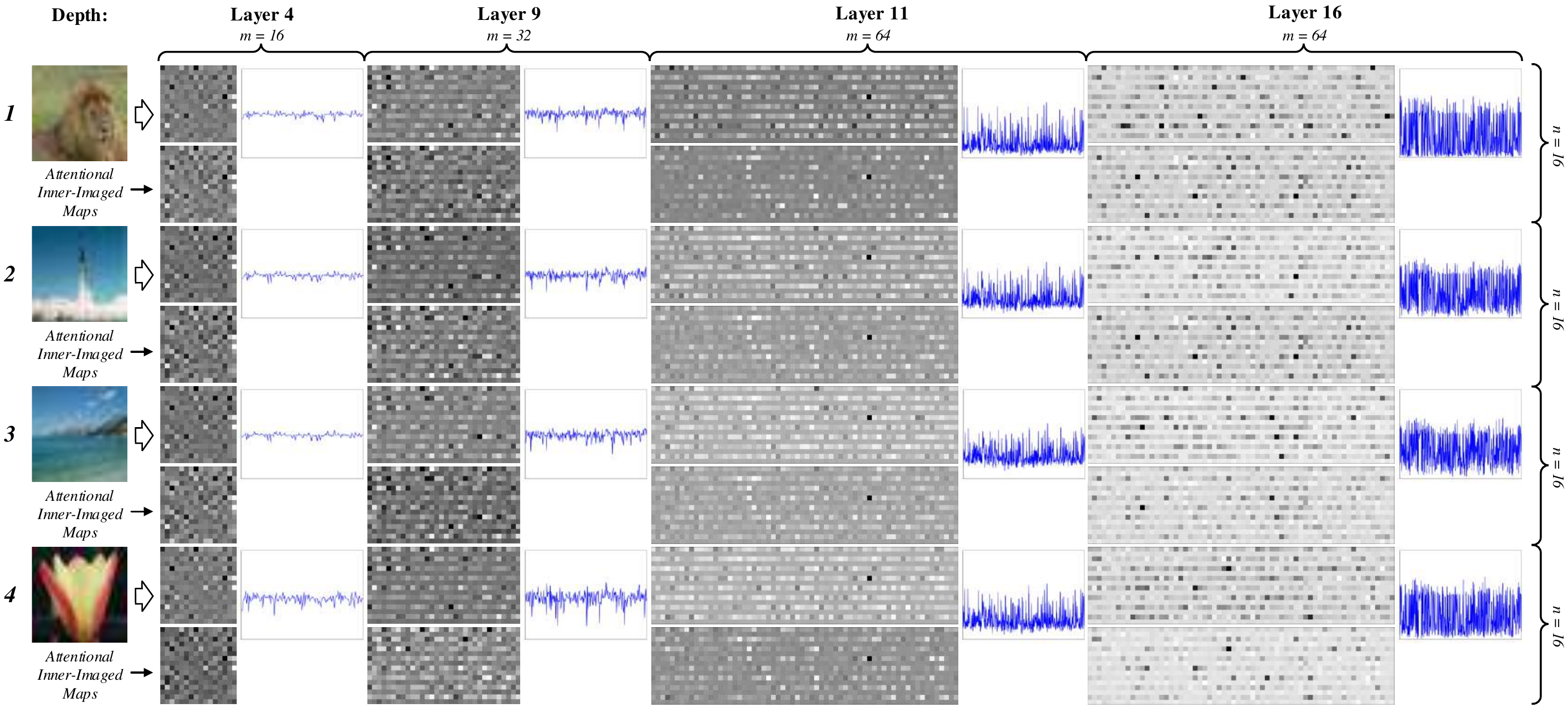}
\caption{Examples of inner-imaged maps and excitation outputs on model: \textbf{CMPE-SE-WRN-16-8}, the first line of each sample represents the initial inner-imaged maps and the next line refers the simulated inner-imaged maps after been re-weighted.}
\label{apx_fig3}
\end{figure*}
\begin{figure*}[t]
\centering
\includegraphics[scale=0.6]{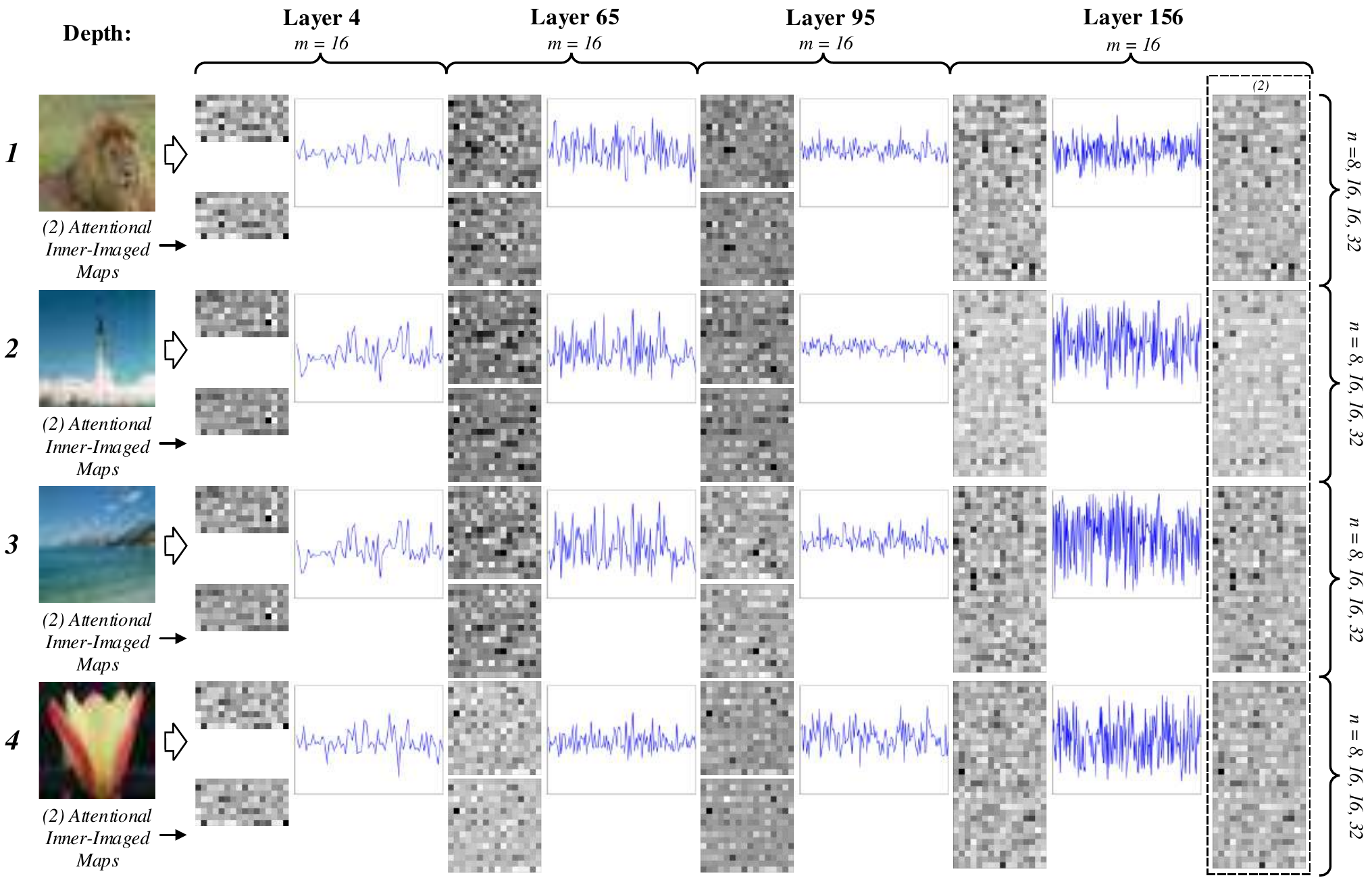}
\caption{Examples of inner-imaged maps and excitation outputs on model: \textbf{CMPE-SE-ResNet-164}, the first line of each sample represents the initial inner-imaged maps and the next line (or with label: '(2)') refers the simulated inner-imaged maps after been re-weighted.}
\label{apx_fig4}
\end{figure*}

Fig \ref{apx_fig1} - \ref{apx_fig4} show the following information: (1) the folded inner-imaged maps before excitation and the simulated inner-imaged maps after been multiplied by the attention value; (2) the channel-wise attentional values. All aforementioned information is taken from the feed-forward results of 4 CIFAR-100 test samples on our models (the same below), which are from 4 layers of different depths, the models contain: CMPE-SE-WRN-28-10, CMPE-SE-WRN-22-10, CMPE-SE-WRN-16-8 and ResNet-164, with folded inner-imaging and \textit{Conv 3$\times$3} pair-view channel relationship encoder.

From these diagrams, we can observe different phenomena for different scale models. Firstly, after the re-weighting with channel-wise attention, all inner-imaged maps have changed compared with the previous ones, and the degree of change depends on the attentional values shown in the diagrams, high fluctuation of attention values can lead to dramatic changes in inner-imaging maps, on the contrary, it will lead to smaller changes or just the difference in color depth.

In most cases of model folded inner-imaging CMPE-SE-WRN-22-10, CMPE-SE-WRN-16-8 and CMPE-SE-ResNet164, with the deepening of layers, channel-wise attention values show more and more strong diversity, and the fluctuation range is more and more intense. In case of CMPE-SE-WRN-28-10, the attention outputs of last layer tend to be stable at near 0.5, with only a few jumps. We infer that in the deeper layer of the high-parameter networks, the feature maps have a strong diversity and high representation ability, so our CMPE-SE module is more inclined to maintain their original information, before that, the CMPE-SE mechanism uses less severe shake for the original features from lower layers, and for the deeper layers of abstract features, more violent shake is automatically applied to enhance the diversity of representation.

For the two-stage inner-imaging samples, we can find that the similarity of inner-imaged maps is very high before being processed by CMPE-SE, for different samples in the same layers, and after channel weight re-scaling, they show a certain degree of differentiation, even if only half of the signals are likely to be re-weighted (signals from identity mappings will not change).

\section{Examples of Excitation Outputs for All Models}
\begin{figure*}[t]
\centering
\includegraphics[scale=0.6]{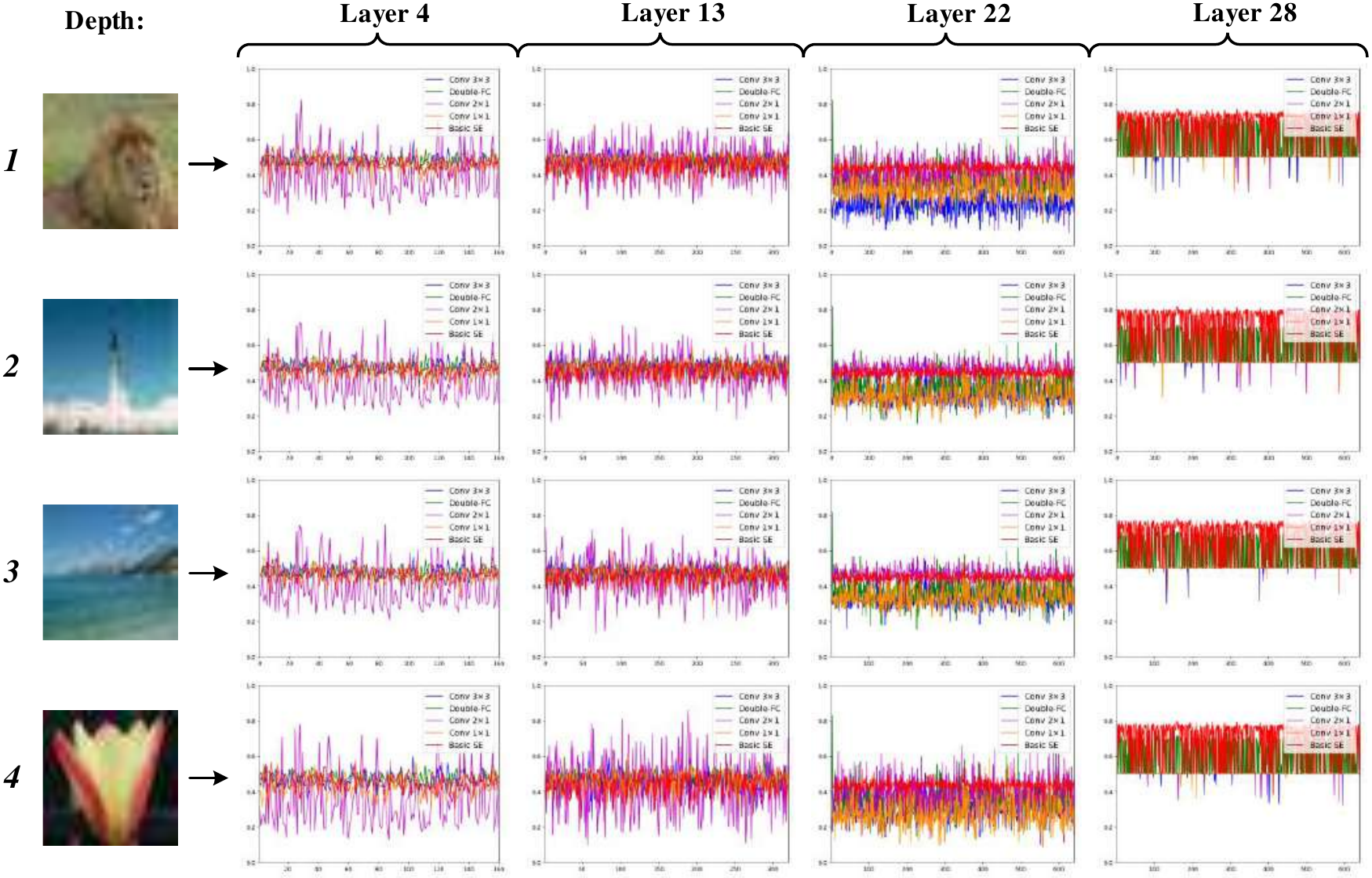}
\caption{Comparison examples of channel-wise attention outputs with different CMPE-SE units and basic SE unit on network: \textbf{WRN-28-10}.}
\label{apx_fig5}
\end{figure*}
\begin{figure*}[t]
\centering
\includegraphics[scale=0.6]{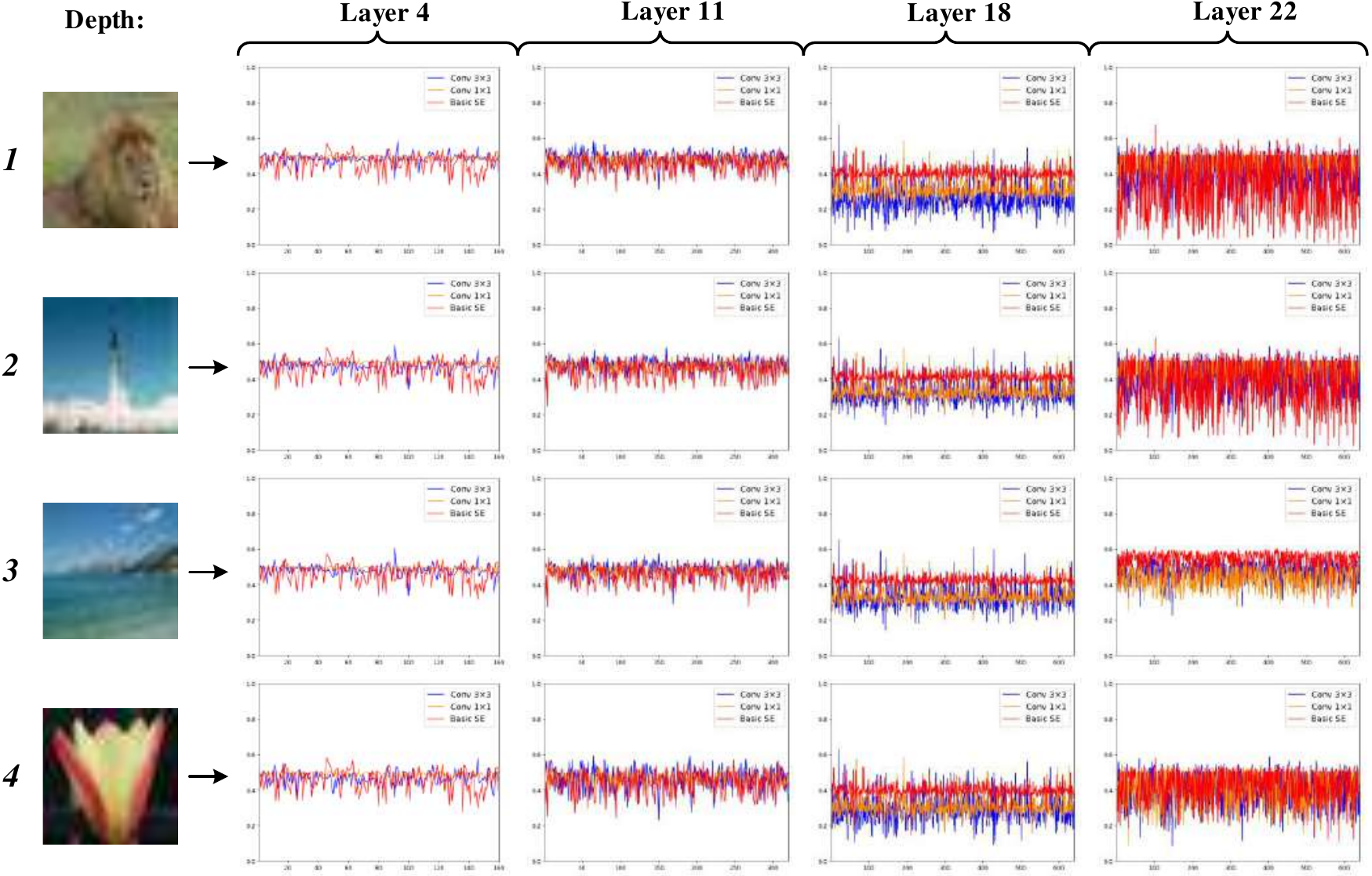}
\caption{Comparison examples of channel-wise attention outputs with different CMPE-SE units and basic SE unit on network: \textbf{WRN-22-10}.}
\label{apx_fig6}
\end{figure*}
\begin{figure*}[t]
\centering
\includegraphics[scale=0.6]{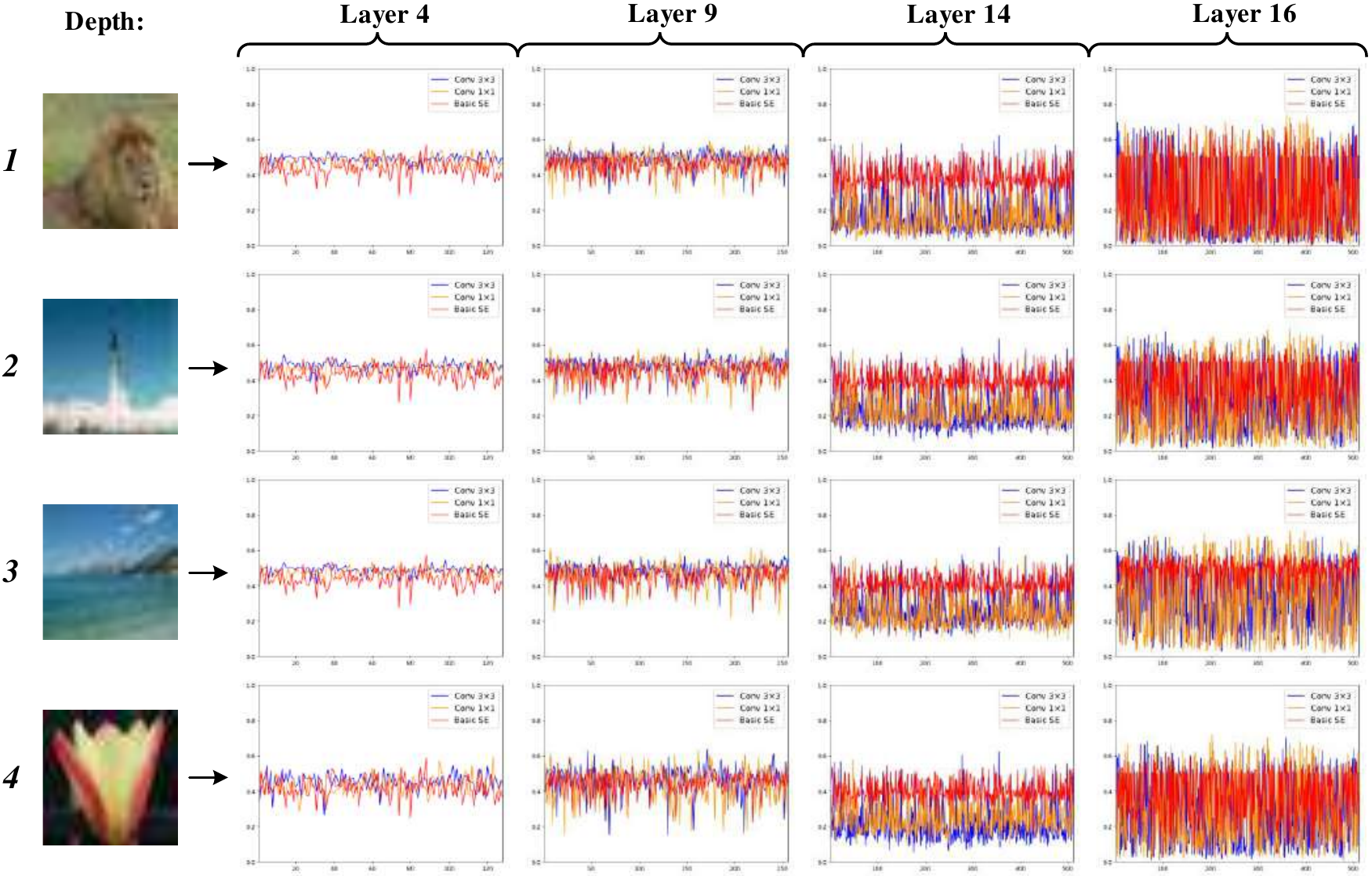}
\caption{Comparison examples of channel-wise attention outputs with different CMPE-SE units and basic SE unit on network: \textbf{WRN-16-8}.}
\label{apx_fig7}
\end{figure*}
\begin{figure*}[t]
\centering
\includegraphics[scale=0.6]{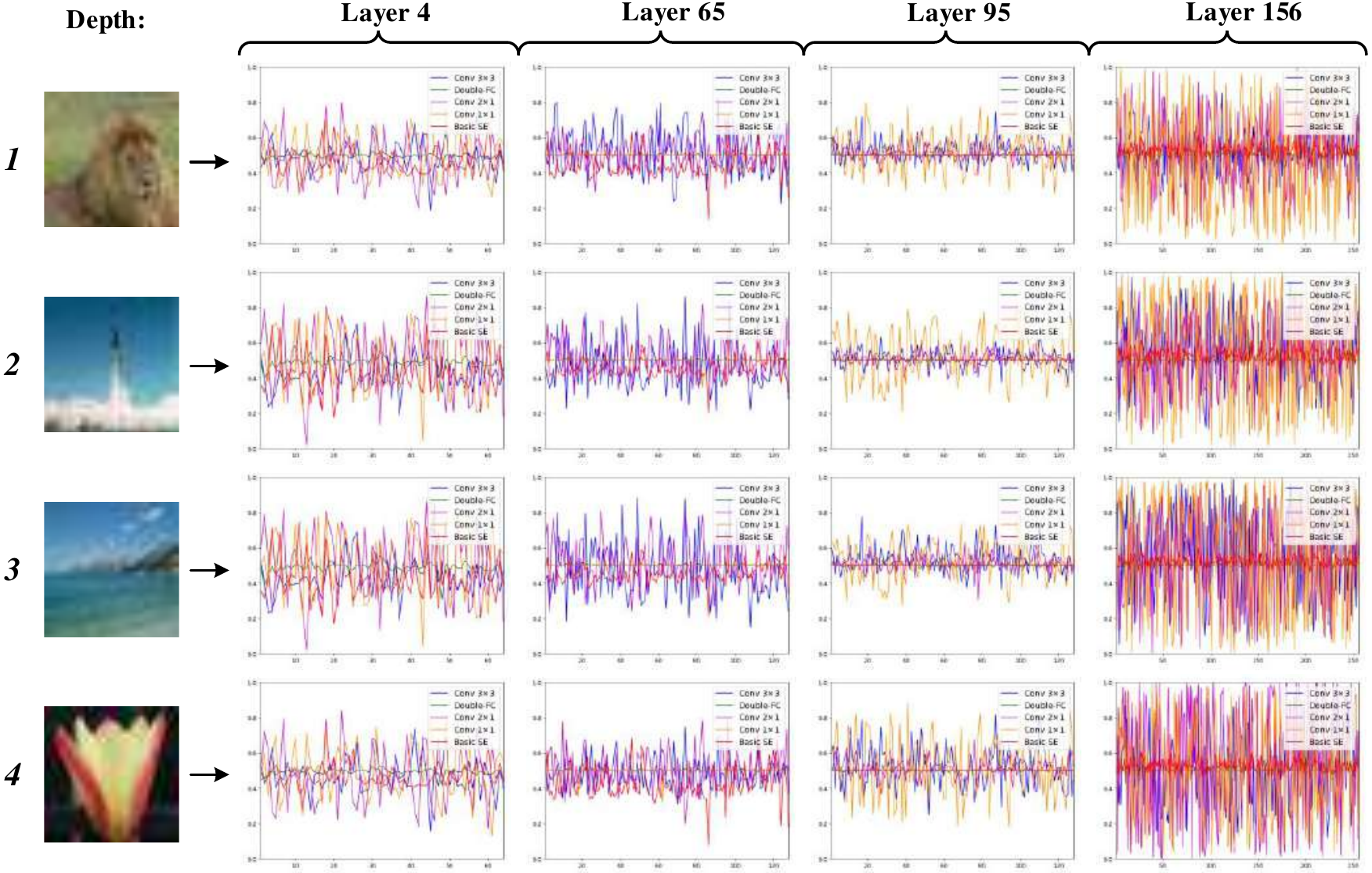}
\caption{Comparison examples of channel-wise attention outputs with different CMPE-SE units and basic SE unit on network: \textbf{ResNet-164}.}
\label{apx_fig8}
\end{figure*}
\begin{figure*}[t]
\centering
\includegraphics[scale=0.6]{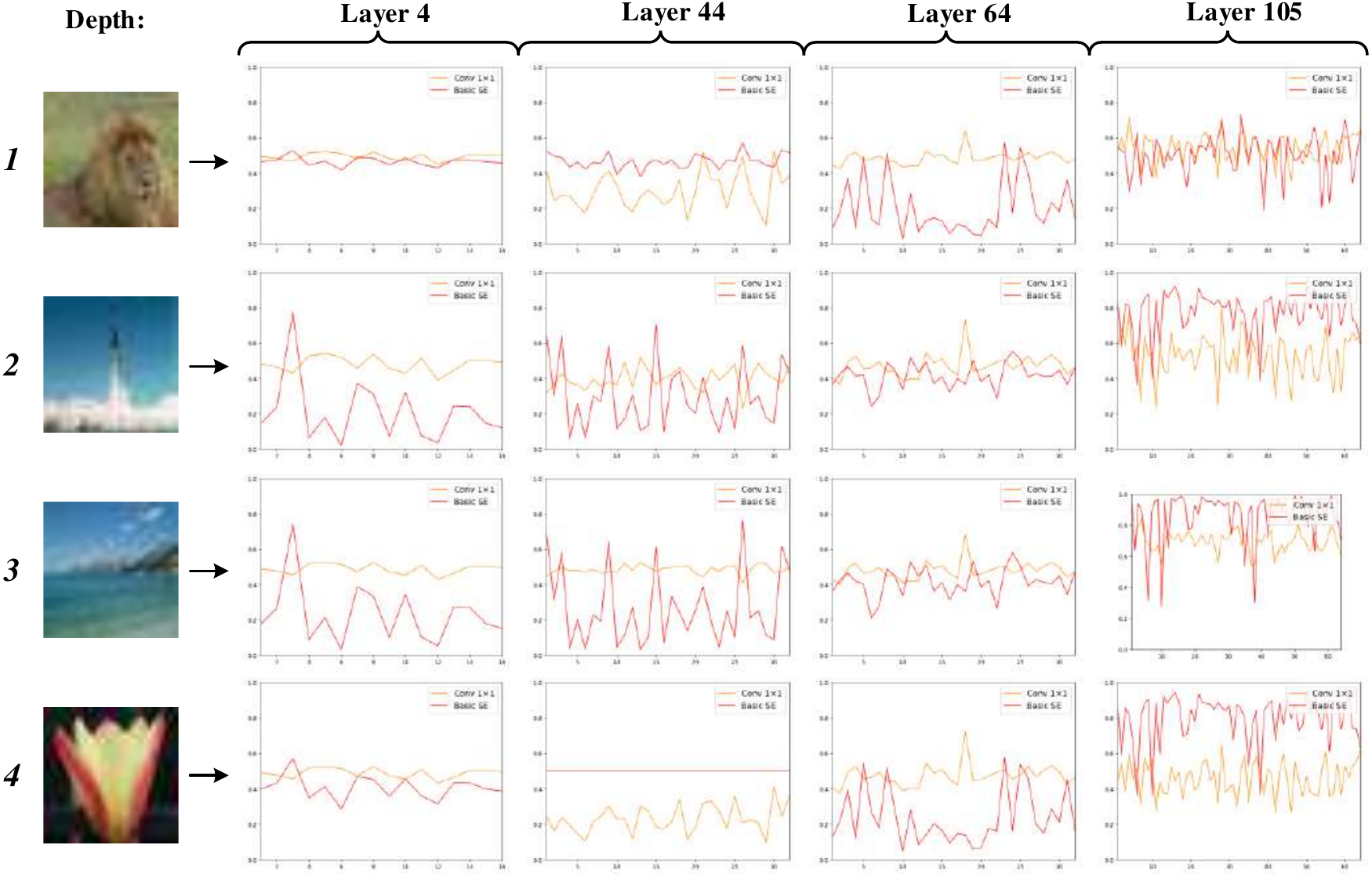}
\caption{Comparison examples of channel-wise attention outputs with different CMPE-SE units and basic SE unit on network: \textbf{ResNet-110}.}
\label{apx_fig9}
\end{figure*}

Fig \ref{apx_fig5} - \ref{apx_fig9} show the comparison results of channel-wise attention by models: WRN-28-10, WRN-22-10, WRN-16-8, ResNet-164 and ResNet-110, with different modes of CMPE-SE and basic SE blocks. For different scale models, attention values from different CMPE-SE modules show different situations.

For model WRN-28-10, WRN-22-10 and WRN-16-8, in deeper layers, attentional excitation values by inner-imaging CMPE-SE modules obviously lower than the outputs of ordinary SE-block, which is represented by the red lines. This phenomenon confirms the following characteristics of CMPE-SE module (especially in modes of inner-imaging): in the deeper layers, the network has modeled some complete features, so the CMPE-SE module will play a role in suppressing redundant residual modeling.

For some networks with more number of layers, like ResNet-164 and ResNet-110, with the deepening of layers, the excitation outputs of basic SE module gradually tend to be flat, while that of CMPE-SE becomes more active.

On the whole, the inner-imaging modes (especially with \textit{Conv 3$\times$3} pair-view encoder) of CMPE-SE module are the most active on all networks. In some ways, this also indicates that the inner-imaging CMPE-SE module works well.

\section{Statistical Analysis of Sample Excitation Outputs}
\begin{figure*}[t]
\centering
\includegraphics[scale=0.6]{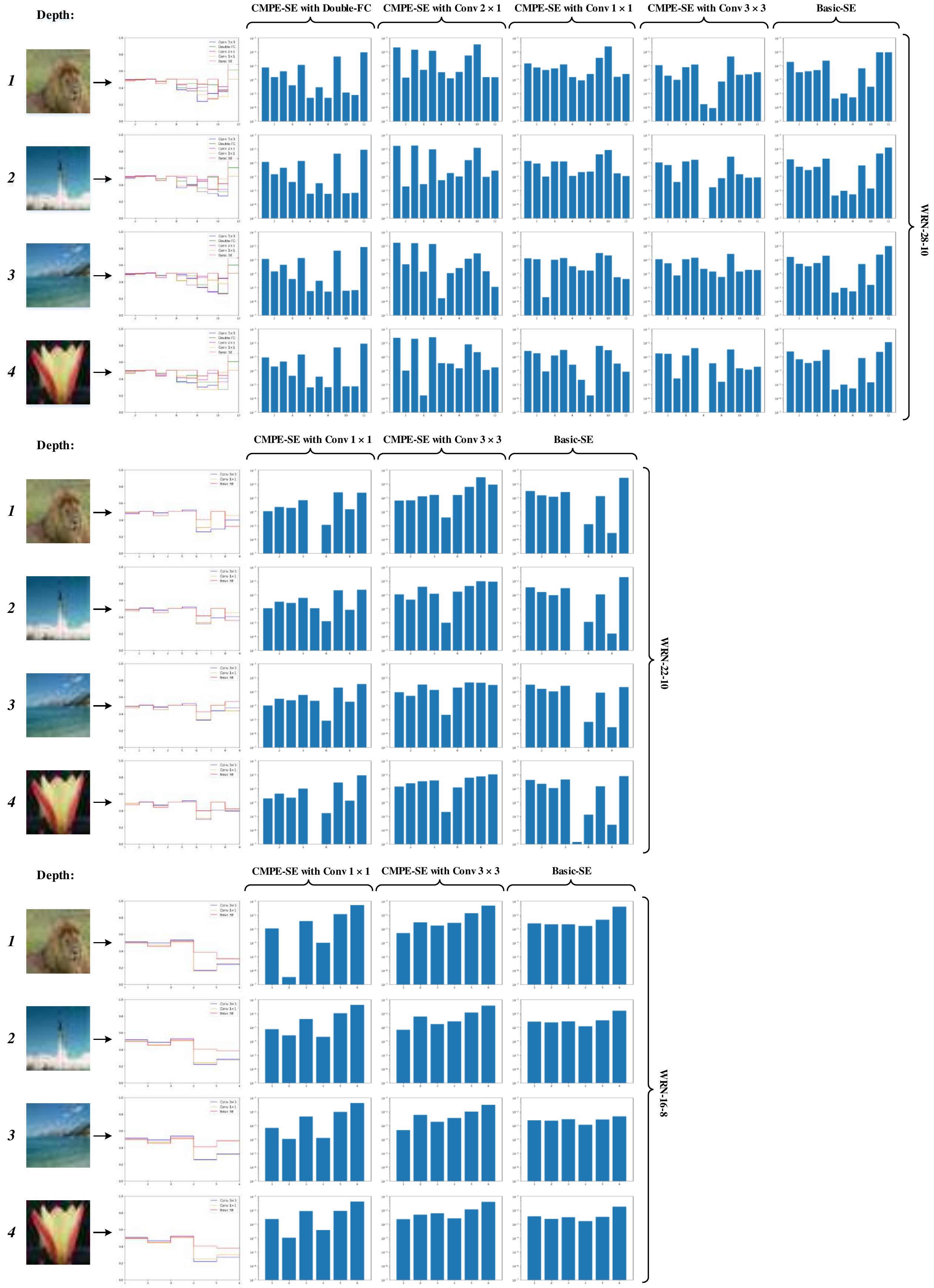}
\caption{Statistical results on examples of channel-wise attention outputs, based on \textbf{wide residual networks}, the ladder diagrams on left refer to the average excitation outputs with different SE units on each block and the bar histograms on the right side show the variances of excitation outputs.}
\label{apx_fig10}
\end{figure*}
\begin{figure*}[t]
\centering
\includegraphics[scale=0.6]{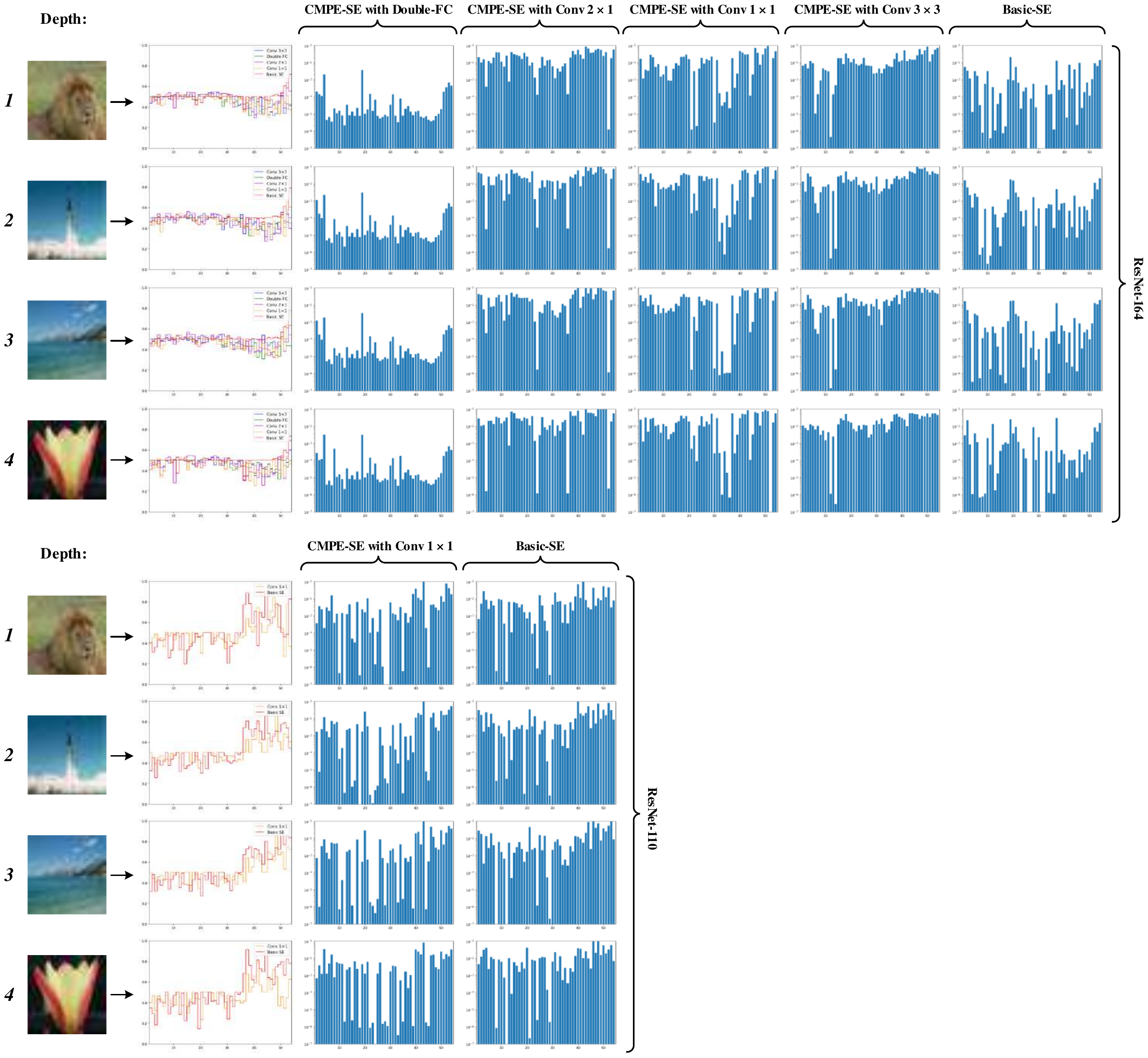}
\caption{Statistical results on examples of channel-wise attention outputs, based on \textbf{pre-act residual networks}, the ladder diagrams on left refer to the average excitation outputs with different SE units on each block and the bar histograms on the right side show the variances of excitation outputs.}
\label{apx_fig11}
\end{figure*}

Furthermore, we exhibit some statistical results of the attentional outputs of each model, in Fig \ref{apx_fig10} and \ref{apx_fig11}. we can also observe some interesting phenomena from them.

The right part of these diagrams show the average attention values of different blocks, firstly, we need note that the number of blocks in networks: WRN-28-10, WRN-22-10, WRN-16-8, ResNet-164 and ResNet-110 are 12, 9, 6, 54 and 54 respectively. In almost all networks, the CMPE-SE module has a obvious inhibition on the residual mappings in the middle and very deep layers, by attentional excitation values, this indicates that the CMPE-SE mechanism does encourage identity mappings at the deeper layers, while reducing the redundancy of residual mapping modeling, which is compared with the basic SE mechanism.

The left part of Fig \ref{apx_fig10} and \ref{apx_fig11} show the variance distributions of attentional outputs with different kinds of SE blocks, which reflect the diversity of channel-wise attention values at each layer. We notice that the variance distributions of excitation outputs of some SE modules are very similar for different samples, on WRN networks, which shows that networks have some similarity in feature weight control of each layer to different images, and some inner-imaging CMPE-SE modules are still able to maintain a few difference on each test samples. For ResNet-164 and ResNet-110, due to the large number of blocks, the distribution of attention value variance shows some variousness, among different samples.

On some blocks of ResNet-164, the variance of attention values by basic SE module becomes very low, representing the corresponding attentional outputs is very flat, while that of the inner-imaging CMPE-SE unit keep at high value in most cases, such as the folded inner-imaging mode with \textit{Conv 3$\times$3} encoder.

\end{document}